\documentclass[final,preprint]{elsarticle}
\usepackage{lineno}
\usepackage[hidelinks]{hyperref}
\modulolinenumbers[5]

\journal{Applied Soft Computing}

\usepackage{amsmath}
\usepackage{amssymb}
\usepackage{amsfonts}

\usepackage{booktabs}
\usepackage{multirow}
\usepackage{subfig}
\usepackage{floatflt}
\usepackage{wrapfig}
\usepackage[linesnumbered, ruled,vlined]{algorithm2e}
\usepackage{url}
\usepackage{xcolor}
\usepackage{adjustbox}
\usepackage{tablefootnote}
\usepackage{lscape}
\usepackage{pdflscape}

\def\vector#1{\mbox{\boldmath $#1$}}

\begin{document}

\begin{frontmatter}

\title{An Analysis of Control Parameters of MOEA/D Under Two Different Optimization Scenarios}


\author{Ryoji Tanabe}
\ead{rt.ryoji.tanabe@gmail.com}

\author{Hisao Ishibuchi\corref{cor1}}
\ead{hisao@sustc.edu.cn}
\cortext[cor1]{Corresponding author}

\address{Shenzhen Key Laboratory of Computational Intelligence, Department of Computer Science and Engineering, Southern University of Science and Technology, Shenzhen, 518055, China}





\begin{abstract}
An unbounded external archive (UEA), which stores all nondominated solutions found during the search process, is frequently used to evaluate the performance of multi-objective evolutionary algorithms (MOEAs) in recent studies.
A recent benchmarking study also shows that decomposition-based MOEA (MOEA/D) is competitive with state-of-the-art MOEAs when the UEA is incorporated into MOEA/D.
However, a parameter study of MOEA/D using the UEA has not yet been performed.
Thus, it is unclear how control parameter settings influence the performance of MOEA/D with the UEA.
In this paper, we present an analysis of control parameters of MOEA/D under two performance evaluation scenarios. One is a final population scenario where the performance assessment of MOEAs is performed based on all nondominated solutions in the final population, and the other is a reduced UEA scenario where it is based on a pre-specified number of selected nondominated solutions from the UEA.
Control parameters of MOEA/D investigated in this paper include the population size, scalarizing functions, and the penalty parameter of the penalty-based boundary intersection (PBI) function.
Experimental results indicate that suitable settings of the three control parameters significantly depend on the choice of an optimization scenario.
We also analyze the reason why the best parameter setting is totally different for each scenario.
\end{abstract}

\begin{keyword}
multi-objective optimization, decomposition based evolutionary algorithms, parameter study, external archive
\end{keyword}

\end{frontmatter}

\nolinenumbers

\section{Introduction}
\label{sec:introduction}

An unconstrained (bound-constrained) multi-objective optimization problem (MOP) can be formulated as follows: 
\begin{align}
\label{eqn:mops}
\text{minimize  } \:\: &\vector{f}(\vector{x}) = \bigl(f_1 (\vector{x}), ..., f_M(\vector{x}) \bigr)^{\rm T}\\
\text{subject to  } &\vector{x} \in \mathbb{S} \subseteq \mathbb{R}^D,\notag 
\end{align}
where $\vector{f}: \mathbb{S} \rightarrow \mathbb{R}^M$ is an objective function vector that consists of $M$ potentially conflicting objective functions, and $\mathbb{R}^M$ is the objective function space.
Here, $\vector{x} = (x_1, ..., x_D)^{\rm T}$ is a $D$-dimensional solution vector, and $\mathbb{S} = \Pi^D_{j=1} [x^{\rm min}_j, x^{\rm max}_j]$ is the bound-constrained search space where $x^{\rm min}_j \leq x_j \leq x^{\rm max}_j$ for each index $j \in \{1, ..., D\}$.



We say that $\vector{x}^1$ dominates $\vector{x}^2$  if and only if $f_i (\vector{x}^1) \leq f_i (\vector{x}^2)$ for all $i \in \{1, ..., M\}$ and $f_i (\vector{x}^1) < f_i (\vector{x}^2)$ for at least one index $i$.
Here, $\vector{x}^*$ is a Pareto-optimal solution if there exists no $\vector{x} \in \mathbb{S}$ such that $\vector{x}$ dominates $ \vector{x}^*$.
 In this case, $\vector{f} (\vector{x}^*)$ is a Pareto-optimal objective function vector.
The set of all $\vector{x}^*$ in $\mathbb{S}$ is the Pareto-optimal solution set (PS), and the set of all $\vector{f}(\vector{x}^*)$ is the Pareto front (PF).
Usually, no solution can simultaneously minimize all objective functions $f_1, ..., f_M$ in MOPs.
Thus, the goal of MOPs is to find a set of nondominated solutions that are well-distributed and close to the PF in the objective function space.
MOPs frequently appear in engineering problems such as aerodynamic wing design problems \cite{OngNK03}, financial and economic problems \cite{PonsichJC13}, oil well problems \cite{SinghRS13}, and unit commitment problems \cite{TrivediSPSR15}.
In general, it is difficult to find a set of good nondominated solutions on MOPs with large values of $M$ and/or $D$. This is because the objective function and solution spaces exponentially grow with $M$ and $D$, respectively.



A multi-objective evolutionary algorithm (MOEA) is an efficient approach for solving MOPs \cite{Deb01}.
Since MOEAs use a set of individuals (solutions of a given MOP) for the search, it is expected that well-distributed nondominated solutions can be found in a single run.
A number of MOEAs have been proposed in the evolutionary computation community \cite{ZhouQLZSZ11,LiLTY15}.
MOEAs can be roughly classified into the following three categories: dominance-based MOEAs, indicator-based MOEAs, and decomposition-based MOEAs.
Dominance-based MOEAs (e.g., NSGA-II \cite{DebAPM02}, SPEA2 \cite{ZitzlerLT01}, and $\epsilon$-MOEA \cite{DebMM05}) mainly use the Pareto dominance or relaxed dominance relations for the mating and environmental selections.
An indicator-based MOEA assigns a so-called fitness value to each individual in the population using a quality indicator \cite{ZitzlerK04}.
Representative indicator-based MOEAs include IBEA \cite{ZitzlerK04}, SMS-EMOA \cite{BeumeNE07}, and HypE \cite{BaderZ11}.
A decomposition-based MOEA decomposes a given MOP with $M$ objectives into multiple single-objective sub-problems\footnote{Some decomposition based MOEAs (e.g., MOEA/D-M2M \cite{LiuGZ14}) decompose a given MOP into a set of simple MOPs.} using a scalarizing function $g: \mathbb{R}^M \rightarrow \mathbb{R}$ and tries to find good solutions for all the subproblems.
Well-known decomposition based MOEAs are MOGLS \cite{IshibuchiM98}, C-MOGA \cite{MurataIG01}, MSOPS \cite{Hughes05}, and MOEA based on decomposition (MOEA/D) \cite{ZhangL07}.
In particular, recent studies report the promising performance of  MOEA/D-type algorithms  \cite{TrivediSSG17}.

MOEA/D \cite{ZhangL07} decomposes an $M$-objective MOP defined in equation \eqref{eqn:mops} into $\mu$ single-objective sub-problems $g_1(\vector{x} | \vector{w}^1), ..., g_{\mu}(\vector{x} | \vector{w}^{\mu})$ using a set of weight vectors $\vector{W} = \{\vector{w}^1, ..., \vector{w}^{\mu}\}$ and a scalarizing function $g$, where $\vector{w}^i = (w^i_1, ..., w^i_M)^{\rm T}$ ($i \in \{1, ..., \mu\}$) and $\sum^M_{j=1} w^i_j = 1$. 
MOEA/D assigns each individual $\vector{x}^i$ ($i \in \{1, ..., \mu\}$) to each sub-problem and tries to find the optimal solution of all subproblems simultaneously.
Unlike other MOEAs, in MOEA/D, the mating and environmental selections are performed only in a set of neighborhood individuals of each weight vector $\vector{w}^i$.
Recent studies show that improved MOEA/D-type algorithms are capable of finding good nondominated solutions on MOPs with complex PFs \cite{LiZ09,JiangY16,YangJJ17,LiuCZD17}.
Although MOEA/D was originally designed for MOPs with up to four objectives, its variants can efficiently handle a large number of objectives \cite{LiDZK15,YuanXWZY16}.
Also, MOEA/D-type algorithms are successfully applied to real-world problems \cite{TrivediSSG17}.

In general, the performance of evolutionary algorithms significantly depends on control parameter settings \cite{EibenHM99,LoboLM07}.
MOEAs including MOEA/D are not an exception.
Therefore, it is important to understand how each control parameter influences the performance of MOEA/D.
General rules of thumb are helpful to users for tuning control parameters of MOEA/D (e.g., the population size $\mu$ should be set to approximately $28$ on three-objective multimodal MOPs).
For these reasons, some parameter studies have been performed  for MOEA/D as briefly reviewed below:

\begin{description}
  \setlength{\itemsep}{-0.4em}
\item [Population size $\mu$:] The population size $\mu$ is an important parameter for all MOEAs.
However, there exist only a few studies that investigate the impact of $\mu$ on the performance of MOEA/D.
Experimental results in the original MOEA/D paper \cite{ZhangL07} show that MOEA/D with a small $\mu$ value ($\mu=20$) can successfully find well-approximated nondominated solutions close to the PF.
The performance of NSGA-II and MOEA/D with various $\mu$ values is investigated in \cite{IshibuchiSTN09}.
Results on multi-objective knapsack problems show that MOEA/D with a very large $\mu$ value works well on most of the problems, while such $\mu$ values make the convergence speed of NSGA-II slow.
In \cite{IshibuchiSMN16a}, the performance of two reference-vector based MOEAs (NSGA-III \cite{DebJ14} and $\theta$-DEA \cite{YuanXWY16}) and two MOEA/D variants (MOEA/D and MOEA/DD \cite{LiDZK15}) with three $\mu$ values are evaluated on three- and five-objective MOPs.
Experimental results in \cite{IshibuchiSMN16a} show that the best $\mu$ value is different for each MOEA.

\item [Scalarizing function $g$:] In MOEA/D, a fitness value of an individual on each sub-problem $j \in \{1, ..., \mu\}$ is given by a pre-defined scalarizing function $g$.
Therefore, the performance of MOEA/D is affected by the scalarizing function $g$ used for the search  \cite{ZhangL07,IshibuchiSTN10,IshibuchiAN15}.
Typical scalarizing functions for MOEA/D include the weighted sum, Chebyshev, and Penalty-based Boundary Intersection (PBI) functions \cite{ZhangL07}.
It is pointed out that MOEA/D with the weighted sum function does not have an ability to handle MOPs with non-convex PFs \cite{ZhangL07}.
In \cite{IshibuchiAN15}, the behavior of MOEA/D with the three representative scalarizing functions is investigated on multi-objective knapsack problems with up to 10 objectives.
Some studies (e.g.,  \cite{ZhangL07,DebJ14}) report that MOEA/D with the PBI function can find evenly distributed nondominated solutions compared to the Chebyshev function.
Since an appropriate scalarizing function is different problem dependent, adaptive selection strategies have also been proposed for these \cite{IshibuchiSTN10,GomezC17}.

\item [Penalty parameter $\theta$:] The efficiency of the PBI function is significantly influenced by the setting of the penalty parameter $\theta$.
The impact of $\theta$ values on the performance of MOEA/D is examined in \cite{MohammadiOLD15,IshibuchiDN16}.
A deterministic control method for $\theta$ is also proposed in \cite{YangJJ17}.

\end{description}

It should be noted that the above-described parameter studies (except for \cite{IshibuchiSMN16a}) are based on only {\em the final population scenario} \cite{TanabeIO17}, where all nondominated solutions in the final population are used for the performance assessment.
The final population scenario is the most widely used optimization scenario and adopted in almost all previous studies of MOEA/D (e.g., \cite{JiangY16,YangJJ17,LiuCZD17,LiDZK15,DebJ14,YuanXWY16}).
While the final population scenario is commonly used in the evolutionary computation community, it is not always a practically desirable optimization scenario \cite{BringmannFK14}.
In the final population scenario, if the number of nondominated solutions found in the search exceeds the predefined population size $\mu$, they are removed from the population to keep the population size constant.
This operation is undesirable if a user wants to know the entire PF using a large number of nondominated solutions.
A good potential solution found in the search is also possibly to be discarded from the population \cite{BringmannFK14}.
Such problems are easily addressed by using an unbounded external archive (UEA) \cite{Hanne99,FieldsendES03} that stores all nondominated solutions found during the search process.
For such reasons, the UEA is frequently used in recent work (e.g., \cite{TanabeIO17,BringmannFK14,WangZILZ17,Lopez-IbanezKL11,RadulescuLS13,BrockhoffTH15}).
The recent COCO platform\footnote{\url{http://coco.gforge.inria.fr/doku.php?id=algorithms-biobj}} with the BBOB-biobj functions \cite{TusarBHA16} also adopts the UEA for the performance assessment of multi-objective optimization methods.

One may think that decision makers usually want to know only a small number of representative, well-distributed nondominated solutions, and thus the use of a large number of nondominated solutions in the UEA for performance assessment does not make sense.
Fortunately, such an issue can be easily addressed by applying a selection method of a small number of nondominated solutions (e.g., \cite{TanabeIO17,BringmannFK14,IshibuchiSTN09b,BasseurDGL16,GuerreiroFP16,ZhangLL09,WangXIWZ17}) to the UEA.
Therefore, there is no particular reason not to incorporate the UEA into MOEAs especially if the computational cost for archive maintenance is sufficiently small in comparison with the objective function evaluation of each solution as in many real-world application problems\footnote{A simulation run that takes a long time is required for some real-world problems to evaluate a single solution \cite{NebroDCLA08,Jin11}.}.
According to \cite{TanabeIO17}, an optimization scenario that uses a pre-specified number of selected nondominated solutions from the UEA for the performance assessment is called {\em the reduced UEA scenario} in this paper.
A benchmarking study in \cite{TanabeIO17} shows that MOEA/D and its variants (MOEA/D-DE \cite{LiZ09} and MOEA/D-DRA \cite{ZhangLL09}) are competitive with state-of-the-art methods under the reduced UEA scenario.



This paper presents an analysis of control parameters of MOEA/D on MOPs with up to five objectives under the final population and reduced UEA scenarios.
Since MOEA/D performs well under the reduced UEA scenario as reported in \cite{TanabeIO17}, such an analysis is worth performing and helpful to both its users and algorithm designers who try to develop more efficient MOEA/D.
We investigate the following three control parameters of MOEA/D: the population size $\mu$, scalarizing functions $g$, and the penalty parameter $\theta$ of the PBI function.
As far as we know, parameter studies of MOEA/D for the reduced UEA scenario have not been well performed.
As mentioned above, most of the previous analytical studies are based only on results for the traditional final population scenario.
Although the reduced UEA scenario is frequently used in recent studies, it is unclear whether MOEA/D with a suitable parameter setting for the final population scenario works similarly under the reduced UEA scenario.
Some MOEA/D algorithms with an external archive have also been proposed (e.g., \cite{ZhangL07,LiDDZ15cec,LiYL16}), but parameter studies are not performed in the literature.
Whereas parameter tuning studies (not analytical studies) of MOEAs for the final population scenario (e.g., \cite{WessingBRN10,BezerraLS17}) and the UEA scenario\footnote{The UEA scenario was named in \cite{TanabeIO17}. All nondominated solutions in the UEA are used for the performance assessment under the UEA scenario.} (e.g., \cite{RadulescuLS13,AnderssonBNS15}) have been presented individually, they do not investigate how different the tuned parameter values are between the two different optimization scenarios.
The main contributions of this paper can be summarized as follows:


\begin{itemize}
  \setlength{\itemsep}{-0.4em}
\item We carefully examine proper settings of the three control parameters ($\mu$, $g$, and $\theta$) which realize the best performance of MOEA/D on the four DTLZ \cite{DebTLZ05} and nine WFG \cite{HubandHBW06} test problems in a component-wise manner.
\item In addition to the one-by-one analysis, we investigate dependencies between two control parameters of MOEA/D.
\item By comparing experimental results between the final population and reduced UEA scenarios, we also reveal how different appropriate settings of the three control parameters of MOEA/D are for each scenario.
\item Furthermore, the reason why the best parameter setting on some MOPs is totally different for each scenario is analyzed in this paper.
\end{itemize}

This paper is organized as follows:
Section \ref{sec:moead} describes the basic procedure of MOEA/D.
Experimental settings are introduced in Section \ref{sec:experimental_settings}.
Section \ref{sec:experimental_results} shows the analysis of the three control parameters ($\mu$, $g$, and $\theta$) of MOEA/D, separately.
Section \ref{sec:further_discussion} presents further analysis of the three control parameters.
Finally, Section \ref{sec:conclusion} concludes this paper and discusses the future work.

\section{MOEA/D}
\label{sec:moead}




Here, MOEA/D \cite{ZhangL07} is briefly explained.
Algorithm \ref{alg:moead07} shows the overall procedure of MOEA/D.
MOEA/D decomposes an MOP with $M$ objectives into $\mu$ single-objective sub-problems using a set of uniformly distributed weight vectors $\vector{W} = \{\vector{w}^1, ..., \vector{w}^{\mu}\}$.
In our study, Das and Dennis's systematic approach \cite{DasD98} was used to generate the weight vectors $\vector{W}$.


At the beginning of the search, all individuals in the population are randomly generated in the search space $\mathbb{S}$ (line 1).
For each subproblem index $i \in \{1, ..., \mu\}$, an index list $\vector{B}^{i} = \{i_1, ..., i_{T}\}$, which is used for the mating and replacement selections, is initialized: $\vector{B}^{i}$ consists of indices of the $T$ closest weight vectors to $\vector{w}^i$ in the weight vector space (lines 2-3) where $T$ is the neighborhood size.

After the initialization, the following steps are repeatedly applied for each subproblem $i \in \{1, .., \mu \}$ until the search termination criteria are met.
For each $i$, the parent indices $k$ and $l$ are randomly selected from $\vector{B}^i$ (line 6).
Then, a child $\vector{u}^i$ is generated by crossing $\vector{x}^{k}$ and $\vector{x}^{l}$ (line 7).
A mutation operator is applied to the child $\vector{u}^i$ if necessary (line 8).
After $\vector{u}^i$ has been generated, the replacement selection is performed using a predefined scalarizing function $g$ (lines 9--11).
For each $j \in \vector{B}^{i}$, the individual $\vector{x}^j$ is compared with the child $\vector{u}^i$ based on $g$.
If $\vector{u}^i$ is better than $\vector{x}^j$ according to their scalarizing function values based on the weight vector $\vector{w}^j$, $\vector{x}^j$ is replaced by $\vector{u}^i$ (lines 10--11).


\IncMargin{0.5em}
\begin{algorithm}[t]
\SetSideCommentRight
$t \leftarrow 1$, initialize the population $\vector{P} =\{ \vector{x}^{1}, ..., \vector{x}^{\mu}\}$\;
\For{$i \in \{1, ..., \mu\}$}{
 Set the neighborhood index list $\vector{B}^{i} = \{i_1, ..., i_{T}\}$\;
}
\While{The termination criteria are not met}{
  \For{$i \in \{1, ..., \mu\}$}{
    Randomly select two indices $k$ and $l$ from $\vector{B}^i$\; 
    Generate a child $\vector{u}^i$ by crossing $\vector{x}^{k}$ and $\vector{x}^{l}$\;
    Apply a mutation operator to $\vector{u}^i$\;
    %
    \For{$j \in \vector{B}^{i}$}{
      \If{$g(\vector{u}^i | \vector{w}^j, \vector{z}^*) \leq g(\vector{x}^j | \vector{w}^j, \vector{z}^*)$} {
        $\vector{x}^{j} \leftarrow \vector{u}^i$\;       
      }
    }
  }
 $t \leftarrow t + 1$\;
}
\caption{The procedure of MOEA/D}
\label{alg:moead07}
\end{algorithm}\DecMargin{0.5em}


Since the replacement of individuals is based on their scalarizing function values, $g$ plays a crucial role in MOEA/D.
Although there are a number of scalarizing functions as reviewed in \cite{Pescador-RojasG17}, in this paper we investigated the following three scalarizing functions: the two Chebyshev functions (a multiplication version $g^{\rm chm}$ \cite{ZhangL07} and a division version $g^{\rm chd}$ \cite{LiZKLW14,QiMLJSW14,YuanXWZY16}) and the PBI function $g^{\rm pbi}$ \cite{ZhangL07}.
Since the three scalarizing functions ($g^{\rm chm}$, $g^{\rm chd}$, and $g^{\rm pbi}$) are most widely used ones in the literature, they are worth investigating.
The three scalarizing functions are defined as follows:
%
\begin{align}
\label{eqn:Chebyshev-mul}
g^{\rm chm}(\vector{x} | \vector{w}, \vector{z}^*) &= \max_{i \in \{1, ..., M\}} \{ w_i |f_i (\vector{x}) - z^*_i|  \},
\\
\label{eqn:Chebyshev-div}
g^{\rm chd}(\vector{x} | \vector{w}, \vector{z}^*) &= \max_{i \in \{1, ..., M\}} \left\{ \frac{|f_i (\vector{x}) - z^*_i|}{w_i}\right\},
\\
\label{eqn:pbi}
g^{\rm pbi}(\vector{x} | \vector{w}, \vector{z}^*) &= d_1 + \theta \, d_2,\\
\label{eqn:pbi_d1}
d_1 &= \frac{\| \left(\vector{f} (\vector{x}) - \vector{z}^* \right)^{\rm T} \, \vector{w}\|}{\|\vector{w}\|},\\
\label{eqn:pbi_d2}
d_2 &= \left\| \vector{f} (\vector{x}) - \left(\vector{z}^* +  d_1 \, \frac{\vector{w}}{\|\vector{w}\|} \right)\right\|,
\end{align}
where all the three scalarizing functions in equations \eqref{eqn:Chebyshev-mul}, \eqref{eqn:Chebyshev-div}, and \eqref{eqn:pbi} should be minimized.
The $\vector{z}^* = (z^*_1, ..., z^*_M)^{\rm T}$ is the ideal point.
Since it is difficult to obtain the actual ideal point $\vector{z}^*$ of a given MOP, its approximated point that consists of the minimum function value for each objective $f_i$ ($i \in \{1, ..., M\}$) found during the search process is usually used for the calculation of equations \eqref{eqn:Chebyshev-mul} -- \eqref{eqn:pbi_d2}.
For $g^{\rm chd}$ in equation \eqref{eqn:Chebyshev-div}, if $w_i = 0$, it was set to $10^{-6}$ for its implementation to avoid division by zero.
While the Chebyshev function $g^{\rm chm}$ defined in equation \eqref{eqn:Chebyshev-mul} is one of the most frequently used scalarizing functions, some studies (e.g., \cite{LiZKLW14,YuanXWZY16}) use its alternative version $g^{\rm chd}$.
This is because the search directions of MOEA/D can be more evenly distributed by using $g^{\rm chd}$ \cite{QiMLJSW14}.

In equations \eqref{eqn:pbi_d1} and \eqref{eqn:pbi_d2}, $d_1$ denotes how close the objective function vector $\vector{f} (\vector{x})$ is to the PF, and $d_2$ is the perpendicular distance between $\vector{f} (\vector{x})$ and $\vector{w}$.
The two distance measures $d_1$ and $d_2$ evaluate the convergence and diversity of the solution $\vector{x}$ in the objective function space, respectively.
The PBI function value calculated by equation \eqref{eqn:pbi} is the sum of $d_1$ and $\theta \,d_2$.
The penalty parameter $\theta > 0 $ controls the balance between the convergence ($d_1$) and diversity ($d_2$) in the population.
While a small $\theta$ encourages convergence toward the PF, a large $\theta$ value emphasizes the importance of diversity in the population \cite{IshibuchiAN15}.




\begin{table}[t]
\begin{center}
  \caption{Properties of the DTLZ and WFG test problems.}
{\scriptsize
  \label{tab:wfg_properties}
\begin{tabular}{llccc}
\midrule
Problem & Shape of PF & Multimodality & Nonseparability & Others\\
\toprule
DTLZ1 & Linear & $\checkmark$ & & \\\midrule
DTLZ2 & Nonconvex &  & & \\\midrule
DTLZ3 & Nonconvex & $\checkmark$ & & \\\midrule
DTLZ4 & Nonconvex & & & Biased\\\midrule
WFG1 & Mixed &  & & Biased\\\midrule
WFG2 & Discontinuous & $\checkmark$ & $\checkmark$ & \\\midrule
\raisebox{0.5em}{WFG3} & \shortstack{Partially\\Degenerate} &  & $\checkmark$ & \\\midrule
WFG4 & Nonconvex & $\checkmark$ &  & \\\midrule
WFG5 & Nonconvex &  &  & Deceptive\\\midrule
WFG6 & Nonconvex &  & $\checkmark$ & \\\midrule
WFG7 & Nonconvex &  &  & Biased\\\midrule
WFG8 & Nonconvex &  & $\checkmark$ & Biased\\\midrule
WFG9 & Nonconvex & $\checkmark$ & $\checkmark$ & Deceptive, Biased\\\midrule
\end{tabular}
}
\end{center}
\end{table}

\section{Experimental settings}
\label{sec:experimental_settings}

This section introduces our experimental settings.
Experimental results are reported in Section \ref{sec:experimental_results} and \ref{sec:further_discussion}.
Subsection \ref{sec:problems} describes test problems and a performance indicator used in our study.
In our study, we used the average performance score (APS) \cite{BaderZ11} to aggregate the performance of MOEA/D with various configurations on 13 MOPs.
The calculation method of the APS value is described in Subsection \ref{sec:aps}.
Subsection \ref{sec:parameter_moead} introduces the parameter settings of MOEA/D.


\subsection{Test problems and performance indicator}
\label{sec:problems}

The four DTLZ \cite{DebTLZ05} and nine WFG  \cite{HubandHBW06} test problems with $M \in \{2, 3, 4, 5\}$ were used in our analysis study.
Table \ref{tab:wfg_properties} summarizes their properties.
The shapes of the PFs of the WFG1, WFG2, and WFG3 test problems are complicated, discontinuous, and partially degenerate \cite{IshibuchiMN16},  respectively.
The DTLZ1 problem has a linear PF.
The shapes of the PFs of the other problems are nonconvex PFs.
According to \cite{DebTLZ05}, for the DTLZ problems, the number of the position variables $k$ was set to $k=5$ for the DTLZ1 problem and $k=10$ for the other  DTLZ problems, where the number of variables is $D=M+k-1$.
Also, as suggested in \cite{HubandHBW06}, for the WFG test problems, the number of the position variables  $k$ was set to $k = 2 \, (M-1)$, and the number of the distance variables $l$ was set to $l=20$, where $D = k+l$.

The hypervolume (HV) indicator \cite{ZitzlerTLFF03} was used for evaluating the quality of a set of obtained nondominated solutions $\vector{A}$.
Before calculating the HV value, the objective function vector $\vector{f} (\vector{x})$ of each $\vector{x} \in \vector{A}$ was normalized using the ideal point and the nadir point of the target MOP.
As suggested in \cite{IshibuchiSMN16}, the reference point for the HV calculation was set to $(1.1, ..., 1.1)^{\rm T}$.
In this setting, the HV range for all of the WFG test problems is $[0, 1.1^M]$.
We further normalized HV values $\in [0, 1.1^M]$ to the range $[0,1]$ by dividing by $1.1^M$.
The HV value of $\vector{A}$ was calculated for every $2 \,000$ function evaluations.

As mentioned in Section \ref{sec:introduction}, all nondominated solutions in the population $\vector{P}$ are used for the HV calculation under the final population scenario.
For the reduced UEA scenario, a selection method of a small number of nondominated solutions from the UEA is necessary.
Although there are some computationally cheap selection methods (e.g., \cite{BringmannFK14,ZhangLL09,WangXIWZ17}), we used a distance-based selection method described in \cite{TanabeIO17}.
In this selection method, a pre-defined number of solutions $b$ is selected from the UEA. 
First,  $M$ extreme solutions having the minimum objective function values for $f_i$ ($i \in \{1, ..., M\}$) are selected for $\vector{B}$.
After that, a nondominated solution which is farthest from a set of already selected ones in the objective function space is repeatedly added to $\vector{B}$ until the size of $\vector{B}$ is equal to $b$.
It is expected that a set of uniformly distributed nondominated solutions in the objective function space are obtained by the distance-based selection method.
In our study, the number of selected nondominated solutions $b$ was set to $200$, $210$, $220$, and $210$ for $M =2$, $3$, $4$, and $5$, respectively.

\begin{figure*}[t]
\small
\newcommand{\widthvar}{0.99}
\centering
\includegraphics[width=\widthvar\textwidth]{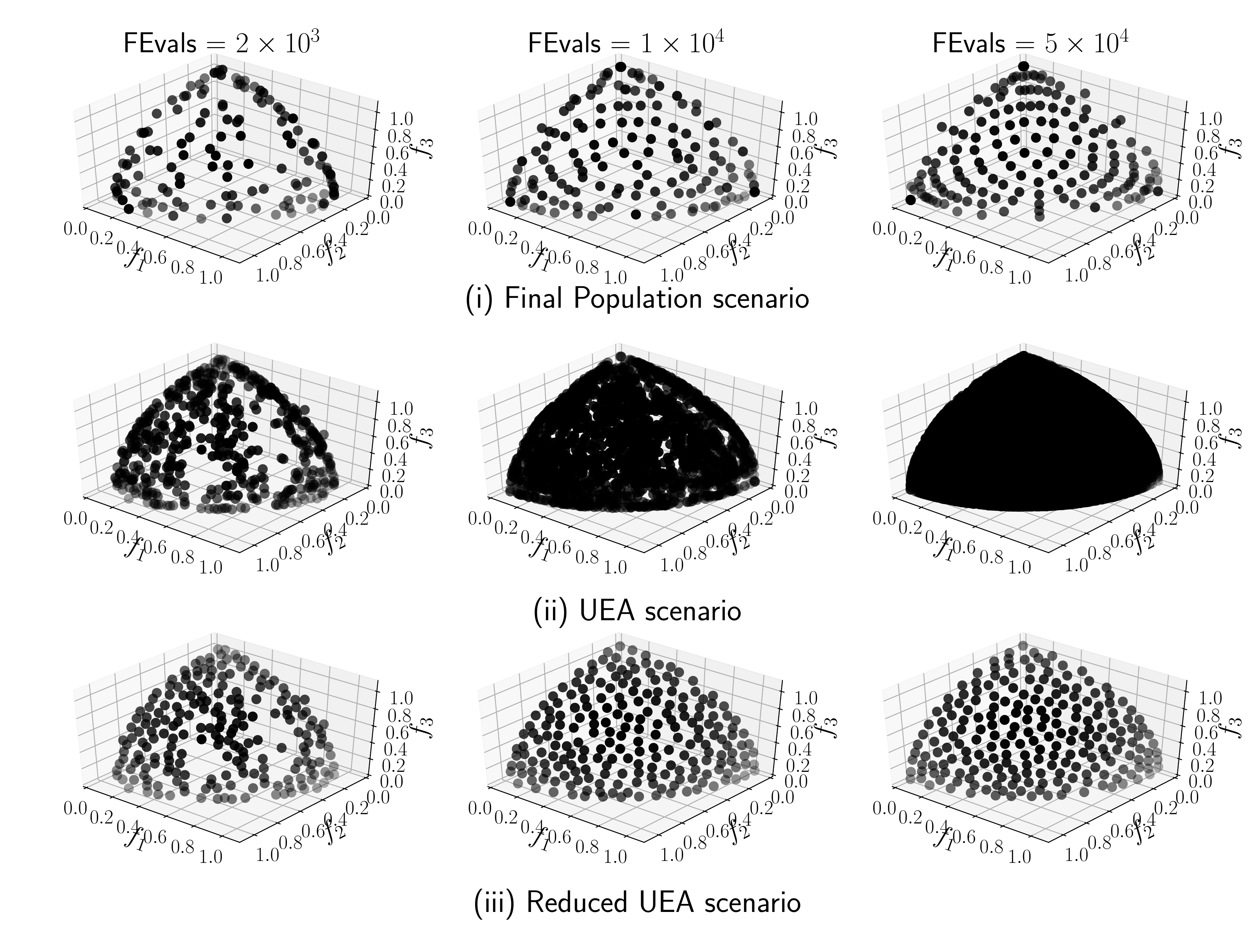}
\caption{Distribution of nondominated solutions found by MOEA/D with $\mu = 210$ and $g^{\rm chm}$ on the three-objective WFG4 problem under (i) the final population scenario, (ii) the UEA scenario, and (iii) the reduced UEA scenario.
Results at $2 \times 10^3$ (left), $1 \times 10^4$ (center), and $5 \times 10^4$ (right) function evaluations (FEvals) are shown.
In the UEA scenario, all nondominated solutions in the UEA are used for the performance assessment \cite{TanabeIO17}.
The number of selected nondominated solutions $b$ for the reduced UEA scenario was set to $210$ for $M = 3$.
Results of a single run are shown.
}
\label{fig:uea_explanation_wfg}
\end{figure*}

For the sake of explanation, we show the distribution of nondominated solutions found by MOEA/D with $\mu = 210$ and $g^{\rm chm}$ on the three-objective WFG4 problem under (i) the final population scenario, (ii) the UEA scenario, and (iii) the reduced UEA scenario in Figure \ref{fig:uea_explanation_wfg}.
Results at $2 \times 10^3$, $1 \times 10^4$, and $5 \times 10^4$ function evaluations are shown.
As shown in Figure \ref{fig:uea_explanation_wfg}, the distribution of nondominated solutions in the population and the UEA is significantly different after $1 \times 10^4$ function evaluations.
In the final population scenario, nondominated solutions to be preserved in the population in the next iteration are determined according to the environmental selection of MOEA/D (lines 9--11 in Algorithm \ref{alg:moead07}).
In contrast, in the UEA scenario, if a newly generated solution is nondominated with respect to existing solutions in the UEA, it enters the UEA.
Then, solutions dominated by the newly inserted solution are discarded from the UEA.
As seen from Figure \ref{fig:uea_explanation_wfg}(ii), the number of nondominated solutions in the UEA gradually increases as the search progresses.
At $5 \times 10^4$ function evaluations, nondominated solutions in the UEA cover the entire PF.
In the reduced UEA scenario, a limited number of nondominated solutions are selected from the UEA using the above-mentioned distance-based selection method.
Sparsely distributed nondominated solutions can be found in Figure \ref{fig:uea_explanation_wfg}(iii).
In summary, the selection of nondominated solutions in the (reduced) UEA scenario is performed independently from the environmental selection of MOEA/D.


\subsection{Average Performance Score (APS)}
\label{sec:aps}

Here, the APS calculation method \cite{BaderZ11} is introduced.
Suppose that $n$ algorithms $A_1, ..., A_n$ are compared for a given problem instance based on the HV values obtained in multiple runs.
For each $i \in \{1, ..., n\}$ and $ j \in \{1, ..., n\} \backslash \{i\}$, let $\delta_{i,j} = 1$, if $A_j$ significantly outperforms $A_i$ using the Wilcoxon rank-sum test with $p < 0.05$, otherwise $\delta_{i,j} = 0$.
Then, the performance score $P(A_i)$ is defined as follows: $P(A_i) = \sum^{n}_{ j \in \{1, ..., n\} \backslash \{i\}} \delta_{i,j}$.
The score $P(A_i)$ represents the number of algorithms outperforming $A_i$. 
The APS value of $A_i$ is the average of the $P(A_i)$ values for all the considered problem instances.
In other words, the APS value of $A_i$ represents how good (relatively) the performance of $A_i$ is among the $n$ algorithms on average over all problem instances. 
A small APS value indicates that the performance of the target algorithm is better than other compared algorithms.

\begin{table*}[t]
\renewcommand{\arraystretch}{1.0}
\begin{center}
  \caption{Settings for the three control parameters of MOEA/D (the population size $\mu$, the scalarizing function $g$, and the penalty value $\theta$ of the PBI function $g^{\rm pbi}$). Unless explicitly noted, the default settings were used in our experimental study. Examined settings in the table denote parameter values which are investigated in our analytical study, and each result can be found in its corresponding section.}
{\small
  \label{tab:moead_parameter_settings}
\scalebox{1.0}[1]{ 
\begin{tabular}{llcccc}
\midrule
\shortstack{Control\\parameters} & \shortstack{Default\\settings} & \shortstack{Examined\\settings} & \shortstack{Corresponding\\sections}\\
\toprule
\shortstack[r]{$\mu$ for $M=2$\textcolor{white}{$\{$}\\$\mu$ for $M=3$\textcolor{white}{$\{$}\\$\mu$ for $M=4$\textcolor{white}{$\{$}\\$\mu$ for $M=5$\textcolor{white}{$\{$}} & \shortstack{$200$\textcolor{white}{$\{$}\\$210$\textcolor{white}{$\{$}\\$220$\textcolor{white}{$\{$}\\$210$\textcolor{white}{$\{$}} & \shortstack[l]{$\{25, 50, 100, 200, 300, 400\}$\\$\{28, 55, 105, 210, 300, 406 \}$\\$\{35, 56, 120, 220, 286, 455 \}$\\$\{126, 210, 330, 495 \}$} &  \raisebox{2em}{Section \ref{sec:results_mu}}\\\midrule
$g$ & $g^{\rm chm}$ & $\{ g^{\rm chm}, g^{\rm chd}, g^{\rm pbi}\}$ & Section \ref{sec:results_g}\\\midrule
$\theta$ & $5$ & $\{0.1, 0.5, 1, 2, 3, 5, 8, 10\}$ & Section \ref{sec:results_pbi_theta}\\\midrule
\end{tabular}
}
}
\end{center}
\end{table*}

\subsection{Parameter settings for MOEA/D}
\label{sec:parameter_moead}


Table \ref{tab:moead_parameter_settings} shows the settings of the three control parameters ($\mu$, $g$, and $\theta$) of MOEA/D.
Unless explicitly noted, the default settings in Table \ref{tab:moead_parameter_settings} were used in our experimental study.
Then, the influence from each control parameter on the performance of MOEA/D is investigated in a component-wise manner.
That is, we analyzed the effect of $\mu$, $g$, and $\theta$ on MOEA/D individually.

We implemented MOEA/D by modifying the source code of MOEA/D-DE \cite{LiZ09} which was downloaded from the jMetal website\footnote{\url{http://jmetal.sourceforge.net/}}.
According to \cite{ZhangL07}, the neighborhood size $T$ of MOEA/D was set to $20$.
The SBX crossover and the polynomial mutation were used in MOEA/D. 
The control parameters of the variation operators were set as follows: $p_c = 1$, $\eta_c = 20$, $p_m = 1/D$, and $\eta_m = 20$.
A simple normalization strategy described in \cite{ZhangL07} was introduced into MOEA/D to handle differently scaled objective function values.

\definecolor{rtdarckgray}{gray}{0.7}
\definecolor{rtlightgray}{gray}{0.7}

\section{Experimental results and discussion}
\label{sec:experimental_results}

Here, we report and discuss the experimental results of MOEA/D with various control parameter settings.
See Table \ref{tab:moead_parameter_settings} for the organization of this section.
Subsections \ref{sec:results_mu}, \ref{sec:results_g}, and \ref{sec:results_pbi_theta} are also organized as follows:
First, we summarize the overall performance of MOEA/D with various configurations on the 13 MOPs based on the APS (Subsections \ref{sec:results_mu_aps}, \ref{sec:results_g_aps}, and \ref{sec:results_pbi_theta_aps}).
Then, experimental results on each problem are described (Subsections \ref{sec:results_mu_each_mop}, \ref{sec:results_g_each_problem}, and \ref{sec:results_pbi_theta_each_problem}).
Finally, we discuss and analyze our results of MOEA/D with each control parameter setting (Subsections \ref{sec:results_mu_discussion}, \ref{sec:results_g_discussion}, and \ref{sec:results_pbi_theta_discussion}).

Since some real-world problems require the execution of a simulation that takes a long time to evaluate the solution, the maximum number of functions evaluations is dependent on the user's available time \cite{NebroDCLA08,Jin11}.
For this reason,  a well-performing MOEA should be able to return a set of good nondominated solutions to a user at any time \cite{RadulescuLS13}.
Also, as pointed out in \cite{BrockhoffTH15}, the comparison based on the end-of-the-run results does not provide sufficient information about the performance of MOEAs.
Therefore, we mainly discuss our experimental results based on the anytime performance of MOEA/D, rather than the end-of-the-run results.



On the one hand, when the number of nondominated solutions obtained by MOEA/D exceeds a predefined $\mu$ value, some of them are discarded from the population in order to keep $\mu$ constant.
Therefore, the quality of solutions, regarding the HV metric, directly depends on $\mu$ and the environmental selection under the final population scenario.
On the other hand, when using the UEA, all nondominated solutions generated by MOEA/D are stored into the UEA independently from the environmental selection.
Thus, the performance of MOEA/D is indirectly influenced by $\mu$ and the environmental selection under the reduced UEA scenario.
Also, a large number of nondominated solutions can be examined and selected for the HV calculation for the reduced UEA scenario.
Due to this reason, in most cases, the HV value under the reduced UEA scenario is better than that under the final population scenario in our experimental results.
For example, see Figure \ref{fig:mu_wfg4}.

\begin{figure*}[t]
\small
\newcommand{\widthvar}{0.99}
\centering
\includegraphics[width=\widthvar\textwidth]{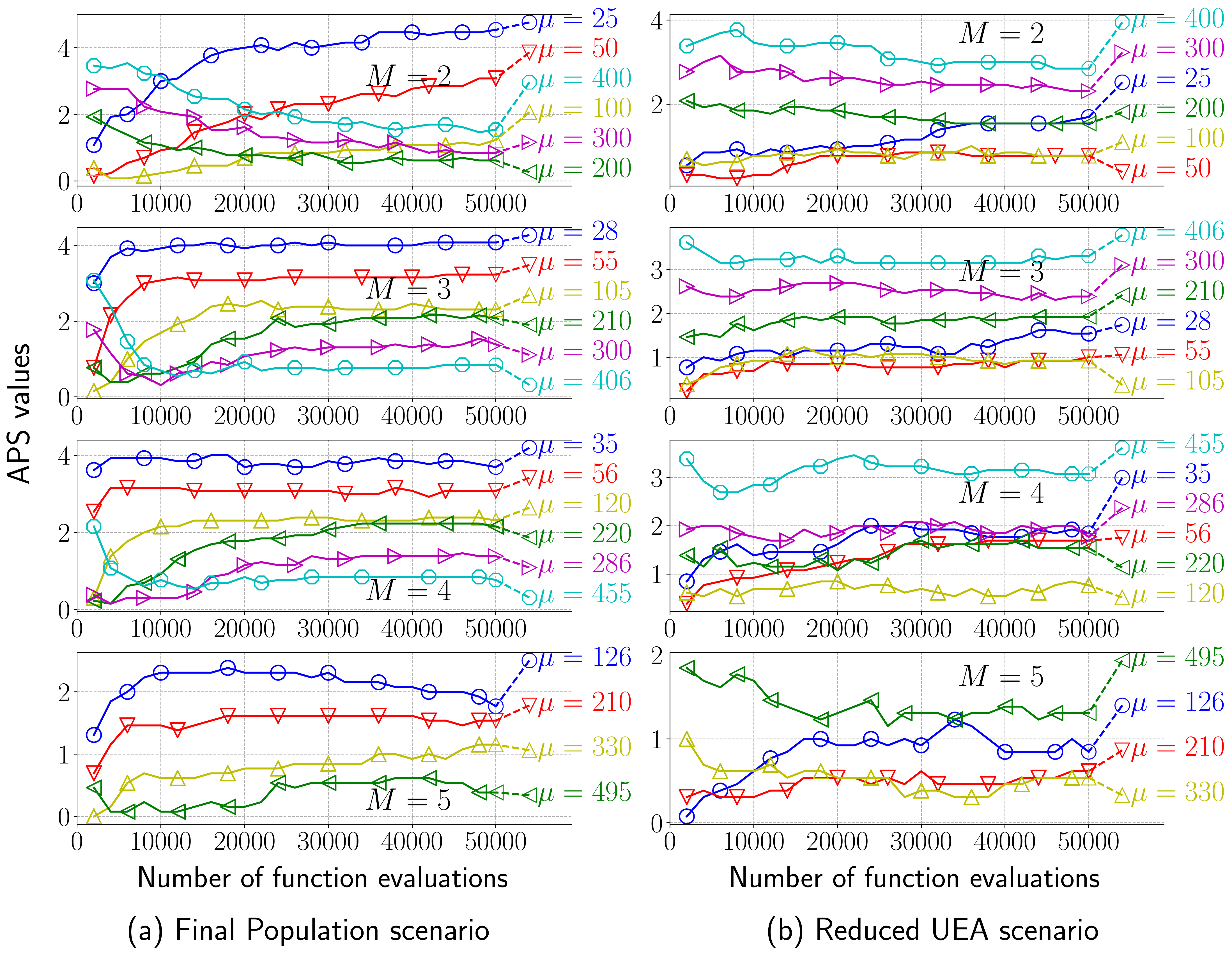}
\caption{
Overall performance of MOEA/D with various $\mu$ settings on the 13 problems (the four DTLZ and nine WFG problems) with $M \in \{2, 3, 4, 5\}$ (lower is better).
The horizontal and vertical axes represent the number of function evaluations and the APS values, respectively.
A small APS value indicates that a corresponding setting performs relatively better than the other settings.
For the APS value \cite{BaderZ11}, see Subsection \ref{sec:aps}.
}
\label{fig:popsize_aps}
\end{figure*}

\subsection{Influence of the population size $\mu$ on the performance of MOEA/D under two scenarios}
\label{sec:results_mu}

\subsubsection{Overall performance}
\label{sec:results_mu_aps}

Figure \ref{fig:popsize_aps} shows the overall performance of MOEA/D with various $\mu$ settings on the 13 problems with $M \in \{2, 3, 4, 5\}$ under  (a) the final population scenario and (b) the reduced UEA scenario.
It should be noted that results for (a) the final population scenario and (b) the reduced UEA scenario shown in this paper are obtained from the identical MOEA/D runs.

First, we describe results of MOEA/D for the final population scenario shown in Figure \ref{fig:popsize_aps}(a).
For $M=2$, while MOEA/D with small $\mu$ values ($\mu \in \{25, 50\}$) perform better than that with larger $\mu$ values ($\mu \in \{100, 200, 300, 400\}$) just after the beginning of the search, their APS values gradually deteriorate as the search progresses.
When $M$ is increasing, such a tendency is more noticeable.
For $M \geq 3$, after several function evaluations, the larger $\mu$ value is used for MOEA/D, the better APS value is achieved.
The best performance of MOEA/D is obtained with the largest $\mu$ value ($\mu = 406, 455,$ and $495$ for $M=3, 4,$ and $5$) for $M \geq 3$, and MOEA/D with the smallest $\mu$ value ($\mu = 25, 28, 35,$ and $126$ for $M=2, 3, 4,$ and $5$) performs worst among all the configurations for all $M$.

Next, the results of MOEA/D for the reduced UEA scenario in Figure \ref{fig:popsize_aps}(b) are described below.
In contrast to the results for the final population scenario, MOEA/D with the largest $\mu$ value always shows the worst APS value.
For $M \in \{2,3\}$, small $\mu$ values ($\mu \in \{50,100\}$ and $\mu \in \{55,105\}$, respectively) are suitable for MOEA/D at any time.
For $M \in \{4,5\}$, MOEA/D with a small $\mu$ value also shows a good performance within a short period.
After several function evaluations, MOEA/D with $\mu=120$ and $\mu \in \{210, 330\}$ outperform other configurations on the four- and five-objective MOPs, respectively.

\begin{figure*}[t]
\small
\newcommand{\widthvar}{0.99}
\centering
\includegraphics[width=\widthvar\textwidth]{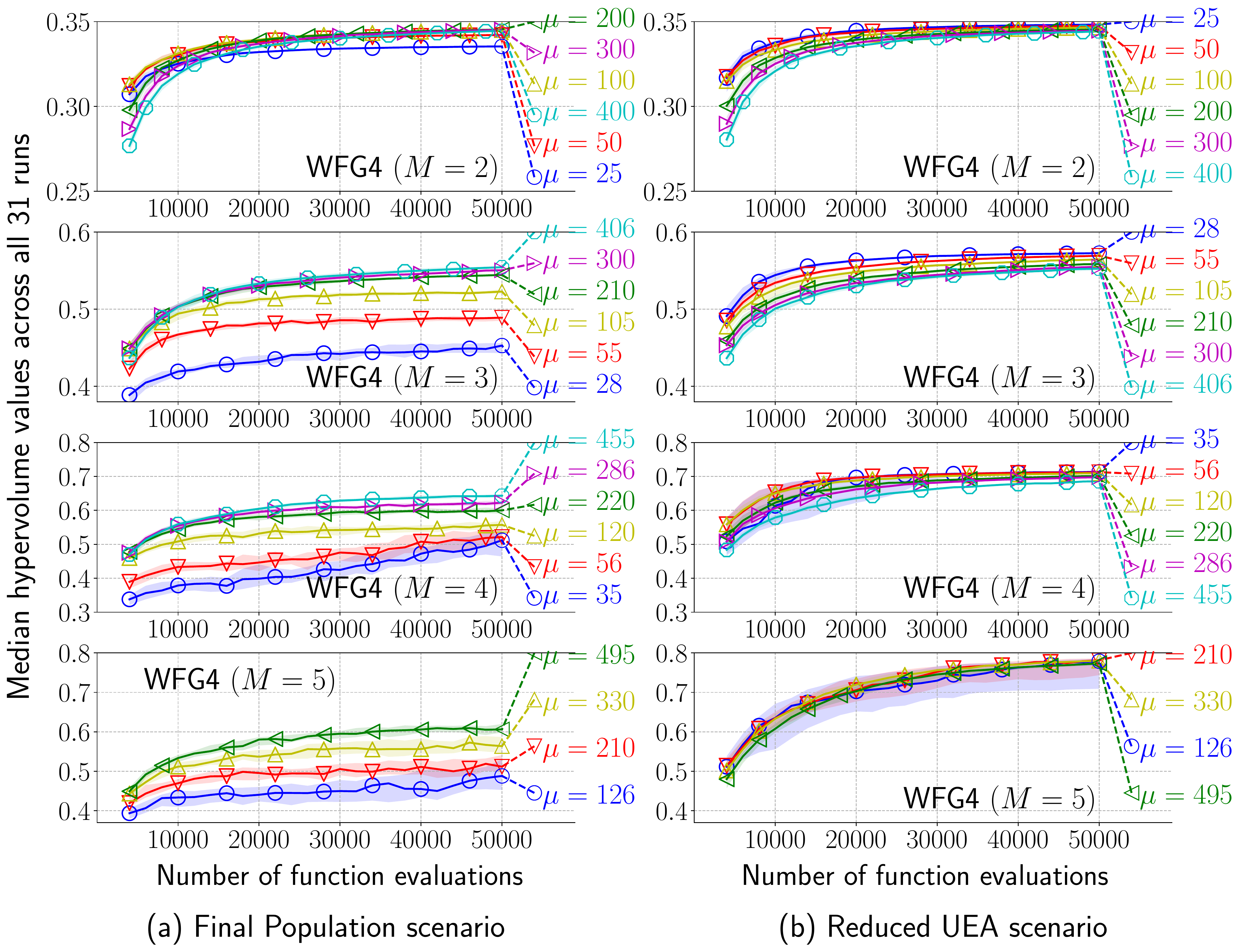}
\caption{
Performance of MOEA/D with various $\mu$ settings on the WFG4 problem with $M \in \{2, 3, 4, 5\}$.
The horizontal and vertical axes represent the number of function evaluations and the HV values, respectively.
The shaded area indicates 25-75 percentiles.
}
\label{fig:mu_wfg4}
\end{figure*}

\subsubsection{Results on each problem}
\label{sec:results_mu_each_mop}

Figure \ref{fig:mu_wfg4} shows the anytime performance of MOEA/D with various $\mu$ settings on the WFG4 problem with $M \in \{2, 3, 4, 5\}$ under two scenarios.
Although we do not show results on all the 13 problems here due to the space constraints, they can be found in Figures S.1 -- S.13 in the supplementary file\footnote{The supplementary file is also available online (\url{https://sites.google.com/site/moeadanalysis/home/ti-moead-asoc17-supp.pdf})} of this paper.

As shown in Figure \ref{fig:mu_wfg4}(a), under the final population scenario, while MOEA/D with $\mu \in \{100, 200\}$ show good performance for $M=2$, MOEA/D with the largest $\mu$ value outperforms other configurations after several function evaluations for $M \geq 3$.
A larger $\mu$ value is beneficial to MOEA/D for the final population scenario. 
Similar results can be found on other problems shown in Figures S.1(a) -- S.13(a).
Of course, there are some exceptions.
For example, results on the WFG1 problem (Figure S.5) show that MOEA/D with a small $\mu$ value exhibits a good performance under the final population scenario.

While good HV values are obtained with a large $\mu$ value in most cases for the final population scenario, MOEA/D with a small $\mu$ value performs the best under the reduced UEA scenario as seen from Figure \ref{fig:mu_wfg4}(b).
It is interesting to notice that the performance rank of MOEA/D with the smallest and largest $\mu$ values is totally different between the two scenarios on the three- and four-objective WFG4 problems.
For example, for $M=3$, while MOEA/D with $\mu=406$ performs significantly better than MOEA/D with $\mu=28$ under the final population scenario, their performance rank is inverted even on the same problem instance in the case of the reduced UEA scenario.
Similar conclusions can be found on most of other problems shown in Figures S.1(b) -- S.13(b), except for the results on the DTLZ4, WFG2, WFG3, and WFG8 problems.
A large $\mu$ value is suitable for MOEA/D on the DTLZ4, WFG2, WFG3, and WFG8 test problems even when we use the reduced UEA scenario.

According to the percentiles in Figure \ref{fig:mu_wfg4}(a) and (b), there is no large variation in the HV values obtained by using all the $\mu$ values for $M=2$.
However, the variation of the HV values becomes large with increasing $M$.
In particular, a larger variation is observed in results of MOEA/D with small $\mu$ values for $M=5$.
Thus, it is likely that the performance of MOEA/D with small $\mu$ values on MOPs with a large $M$ is significantly different for each run.


\subsubsection{Discussion}
\label{sec:results_mu_discussion}



We here discuss why the best population size $\mu$ is dependent on the choice of an optimization scenario, the number of objectives $M$, and the characteristics of a given problem.
As described in Subsections \ref{sec:results_mu_aps} and \ref{sec:results_mu_each_mop}, while MOEA/D with the largest $\mu$ value outperforms other configurations in most cases under the final population scenario, a small $\mu$ value is suitable for MOEA/D on some problems for the reduced UEA scenario.
In particular, the performance rank of MOEA/D with various $\mu$ values on the WFG4 problem with $M \in \{3, 4\}$ is totally different depending on the considered optimization scenario.

This seems to be simply because MOEA/D with a large $\mu$ value is capable of keeping a large number of nondominated solutions in the population.
As discussed in \cite{IshibuchiMN15}, such a large number of nondominated solutions are always beneficial for the HV calculation.
For this reason, the results in Subsections \ref{sec:results_mu_aps} and \ref{sec:results_mu_each_mop} show that MOEA/D with a large $\mu$ value performs well on most problems.
However, nondominated solutions in the population do not directly influence the performance of MOEA/D under the reduced UEA scenario, because solutions selected from the UEA are used for the HV calculation.
Also, MOEA/D can store nondominated solutions whose size is over $\mu$ in the UEA.
Due to this reason, MOEA/D with a small $\mu$ value works well under the reduced UEA scenario.

\begin{figure*}[t]
\small
\newcommand{\widthvar}{0.99}
\centering
\includegraphics[width=\widthvar\textwidth]{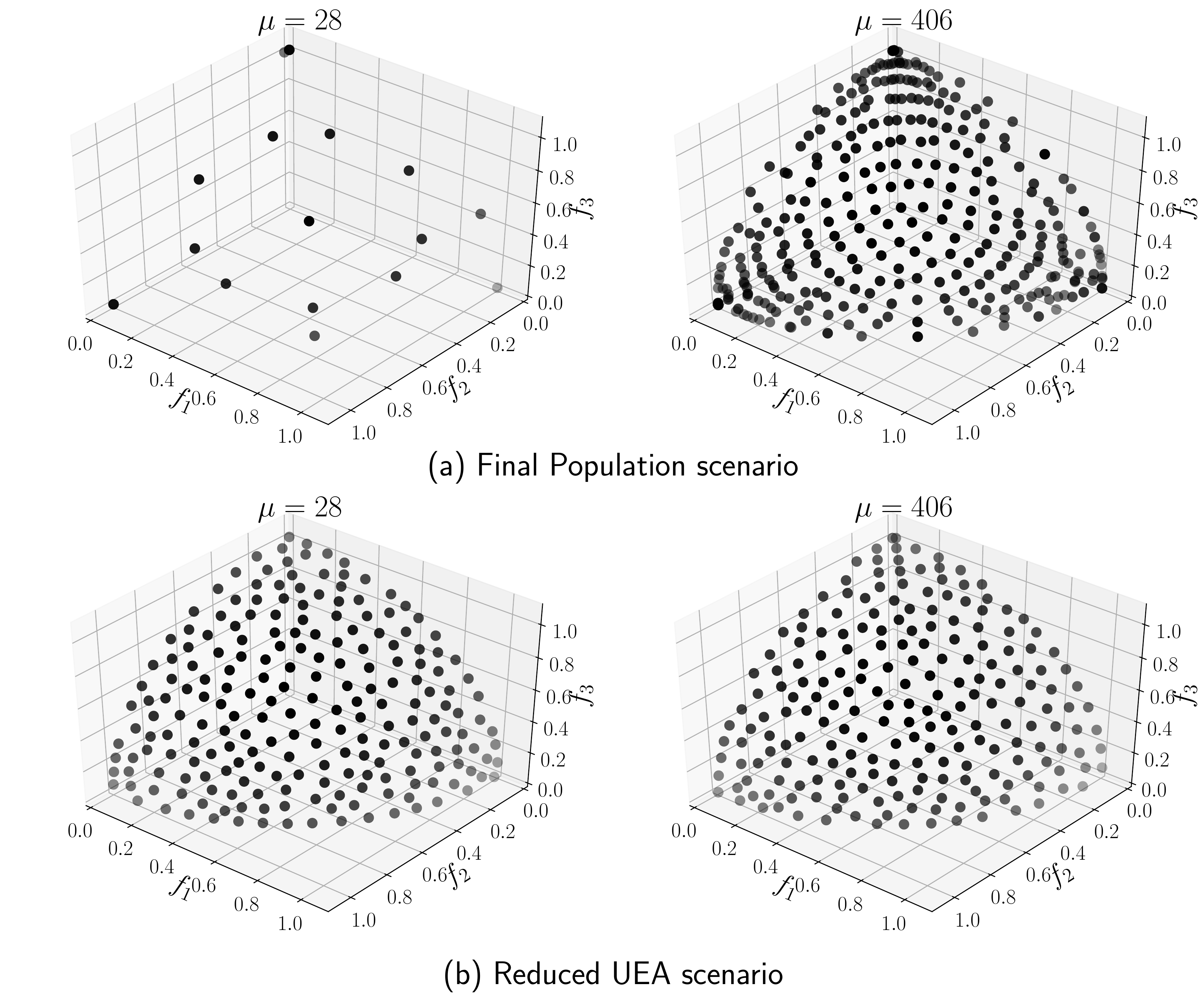}
\caption{
Distribution of nondominated solutions in the objective function space on the three-objective WFG4 problem for the final and reduced UEA scenarios.
Nondominated solutions shown here are found by MOEA/D with $\mu=28$ and $\mu=406$ at $5\times 10^4$ function evaluations.
The number of selected nondominated solutions $b$ for the reduced UEA scenario was set to $210$ for $M = 3$.
Results of a single run with the median HV value are shown.
}
\label{fig:mu_distribution_wfg4}
\end{figure*}

Figure \ref{fig:mu_distribution_wfg4} shows the distribution of nondominated solutions in the objective function space for each scenario found by MOEA/D with $\mu=28$ and $\mu=406$ on the three-objective WFG4 problem.
For the final population scenario, while MOEA/D with $\mu=406$ achieves a set of diverse nondominated solutions, MOEA/D with $\mu=28$ can provide only a small number of nondominated solutions.
Unlike the results for the final population scenario, MOEA/D with both $\mu$ settings successfully maintain the densely distributed solutions covering the entire PF in the UEA  for the reduced UEA scenario.
It is likely that MOEA/D with a small $\mu$ value cannot keep good nondominated solutions in the population but has an ability to generate well-distributed solutions.


We do not intend to argue that a small $\mu$ value is always beneficial for MOEA/D under the reduced UEA scenario.
Our results in Subsection \ref{sec:results_mu_each_mop} show that a (relatively) large $\mu$ value is necessary for MOEA/D to find good nondominated solutions on some problems, especially MOPs with a large $M$ or an irregular shape of the PF.
As seen from the results on the five-objective WFG4 problem (Figure \ref{fig:mu_wfg4}), while MOEA/D with the smallest $\mu$ value ($\mu=126$) achieves the highest HV value until $14\,000$ function evaluations, it clearly performs worst after that.
Since the objective function space is exponentially increasing with $M$, a large population size is necessary for the search if $M$ is large.

Results on the DTLZ4, WFG2, WFG3, and WFG8 problems (Figure S.4, S.6, S.7, and S.12, respectively) show that a large $\mu$ value is beneficial for MOEA/D for all $M$  independently of the choice of an optimization scenario.
The shape of the PF of the WFG2 and WFG3 problems are discontinuous and partially degenerate \cite{IshibuchiMN16}, respectively.
Since the Pareto optimal solutions do not exist in most of the decomposed subproblems of MOPs having discontinuous and degenerate PFs \cite{AsafuddoulaSR17}, a large $\mu$ value, which makes a large number of subproblems, is necessary for MOEA/D to find well-approximated solutions.
For this reason, MOEA/D with a small $\mu$ value performs poorly on the WFG2 and WFG3 problems.
Also, the mapping from the solution space to the objective function space $\vector{f}: \mathbb{S} \rightarrow \mathbb{R}^M$ is biased in the DTLZ4 and WFG8 problems.
The existence of the bias makes the problem more difficult, and thus MOEA/D with a small $\mu$ value cannot find good solutions on the DTLZ4 and WFG8 problems.





\begin{figure*}[t]
\small
\newcommand{\widthvar}{0.99}
\centering
\includegraphics[width=\widthvar\textwidth]{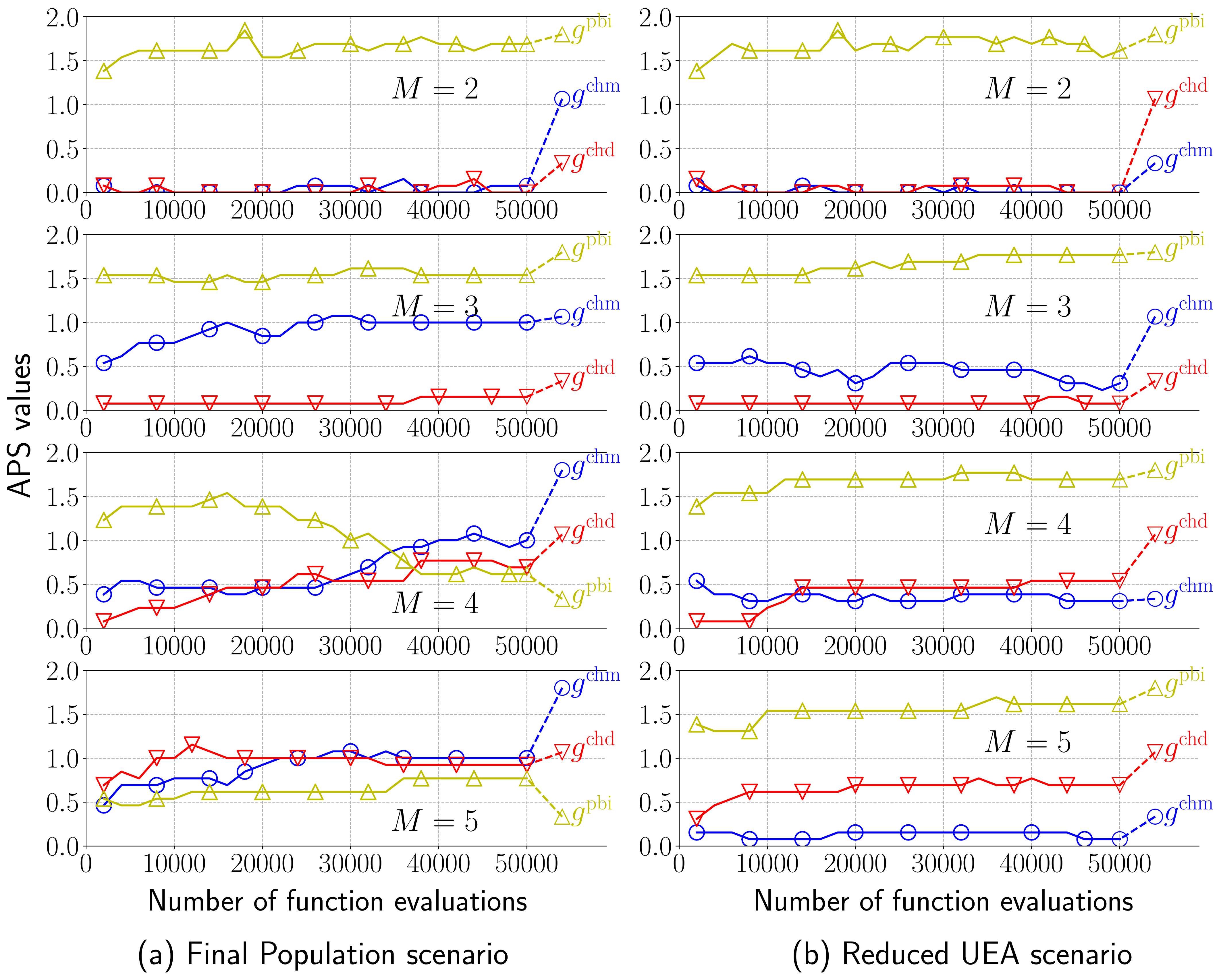}
\caption{
Overall performance of MOEA/D with the three scalarizing functions ($g^{\rm chm}$, $g^{\rm chd}$, and $g^{\rm pbi}$ with $\theta=5$) on the 13 problems (the four DTLZ and nine WFG problems) with $M \in \{2, 3, 4, 5\}$ (lower is better).
The horizontal and vertical axes represent the number of function evaluations and the APS values, respectively.
For the APS value \cite{BaderZ11}, see Subsection \ref{sec:aps}.
}
\label{fig:sf_aps}
\end{figure*}

\begin{figure*}[t]
\small
\newcommand{\widthvar}{0.99}
\centering
\includegraphics[width=\widthvar\textwidth]{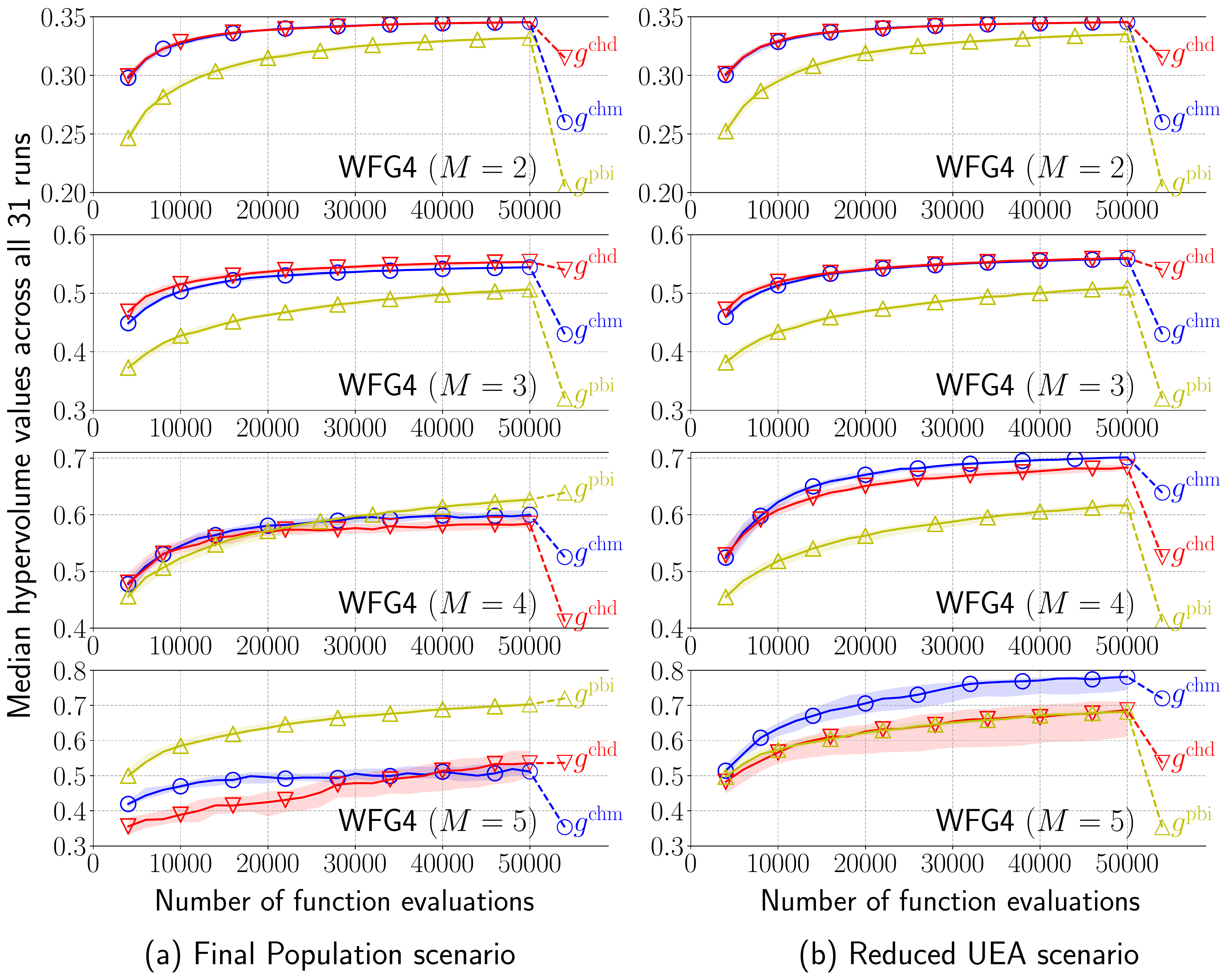}
\caption{
Performance of MOEA/D with the three scalarizing functions ($g^{\rm chm}$, $g^{\rm chd}$, and $g^{\rm pbi}$ with $\theta=5$) on the WFG4 problem with $M \in \{2, 3, 4, 5\}$.
The horizontal and vertical axes represent the number of function evaluations and the HV values, respectively.
The shaded area indicates 25-75 percentiles.
}
\label{fig:sf_wfg4}
\end{figure*}

\subsection{Impact of the scalarizing function $g$ on the performance of MOEA/D under two scenarios}
\label{sec:results_g}

\subsubsection{Overall performance}
\label{sec:results_g_aps}

Figure \ref{fig:sf_aps} shows the overall performance of MOEA/D with the three scalarizing functions ($g^{\rm chm}$, $g^{\rm chd}$, and $g^{\rm pbi}$ with $\theta=5$) over all the 13 test problems. 
For details of the three functions, see Section \ref{sec:moead}.
The influence of $\theta$ on the performance of MOEA/D using $g^{\rm pbi}$ is investigated in Subsection \ref{sec:results_pbi_theta}.

The results on the two- and three-objective 13 problems for the final population scenario show that the performance of MOEA/D with the two Chebyshev functions ($g^{\rm chm}$ and $g^{\rm chd}$) is better than that with $g^{\rm pbi}$ at anytime (Figure \ref{fig:sf_aps}(a)).
For $M=4$, MOEA/D with $g^{\rm pbi}$ outperforms the other configurations after $38\,000$ function evaluations.
For $M=5$, MOEA/D with $g^{\rm pbi}$ always exhibits the best performance.

The results for the reduced UEA scenario (Figure \ref{fig:sf_aps}(b)) are significantly different from the results for the final population scenario described above.
As seen from Figure \ref{fig:sf_aps}(b), MOEA/D with the two Chebyshev functions are competitive with each other. 
The $g^{\rm chm}$ and $g^{\rm chd}$ are the best scalarizing functions for $M \in \{2,4,5\}$ and $M =3$, respectively.
Interestingly, while MOEA/D with $g^{\rm pbi}$ exhibits a good performance on MOPs with more than four objectives after several function evaluations under the final population (Figure \ref{fig:sf_aps}(a)), its performance is worst among the three configurations for all $M$ under the reduced UEA scenario (Figure \ref{fig:sf_aps}(b)).

\subsubsection{Results on each problem}
\label{sec:results_g_each_problem}

Figure \ref{fig:sf_wfg4} shows the anytime performance of MOEA/D with the three scalarizing functions on the WFG4 problem with $M \in \{2, 3, 4, 5\}$.
Results on all the 13 problems can be found in Figures S.14 -- S.26 in the supplementary file.


As shown in Figure \ref{fig:sf_wfg4}, the results  on the WFG4 problem are similar to the aggregated results on all of the 13 problems shown in Figure \ref{fig:sf_aps}.
For both scenarios, MOEA/D with the two Chebyshev functions perform significantly better than that with $g^{\rm pbi}$ on the two- and three-objective WFG4 problems.
For the final population scenario, with increasing $M$, MOEA/D with $g^{\rm pbi}$ achieves the highest HV values among the three configurations in the later search stage.
However, for the reduced UEA scenario, the worst HV values are obtained by using $g^{\rm pbi}$, and MOEA/D with the two Chebyshev functions show better performance at any time.
As seen from the percentiles in Figure \ref{fig:sf_wfg4}(a) and (b), the variation of the HV values achieved by using the two Chebyshev functions is large for $M=5$.
In contrast, the performance of MOEA/D with $g^{\rm pbi}$ is not much different for each run.
The quality of nondominated solutions found by using $g^{\rm pbi}$ is likely to be stable even on MOPs with a large $M$.


Results on the DTLZ2, WFG5, WFG6, WFG7, WFG8, and WFG9 problems are similar to results on the WFG4 problem.
On the DTLZ1 problem, MOEA/D with $g^{\rm chd}$ and $g^{\rm pbi}$ show good performance (Figure S.14).
On the DTLZ3 problem, $g^{\rm chd}$ is the best scalarizing function for all $M$ (Figure S.16).
On the DTLZ4 problem with $M \geq 3$, MOEA/D with $g^{\rm pbi}$ performs the best (Figure S.17).
On the WFG2 and WFG3 problems with all $M$, the performance of MOEA/D with the two Chebyshev functions is better than that with $g^{\rm pbi}$ even for the final population scenario (Figure S.19 and S.20).

\subsubsection{Discussion}
\label{sec:results_g_discussion}


A number of previous work report that MOEA/D with $g^{\rm chm}$ is not capable of obtaining well-distributed nondominated solutions in the objective function space (e.g., \cite{ZhangL07,DebJ14,LiYL16,GomezC15}).
Recent studies also report the promising performance of MOEA/D with $g^{\rm pbi}$ on MOPs with more than four objectives \cite{LiDZK15,DebJ14}.
Our results for the final population described in Subsections \ref{sec:results_g_aps} and \ref{sec:results_g_each_problem} are consistent with these previous studies.
However, the performance of MOEA/D with the two Chebyshev functions ($g^{\rm chm}$ and $g^{\rm chd}$) is significantly better than that with $g^{\rm pbi}$ on most problem instances under the reduced UEA scenario.
In summary, the best scalarizing function for MOEA/D is significantly dependent on the choice of an optimization scenario.
Below, we discuss this result (mainly with respect to the comparison between the PBI function and the two Chebyshev functions).



\begin{figure*}[t]
\small
\newcommand{\widthvar}{0.99}
\centering
\includegraphics[width=\widthvar\textwidth]{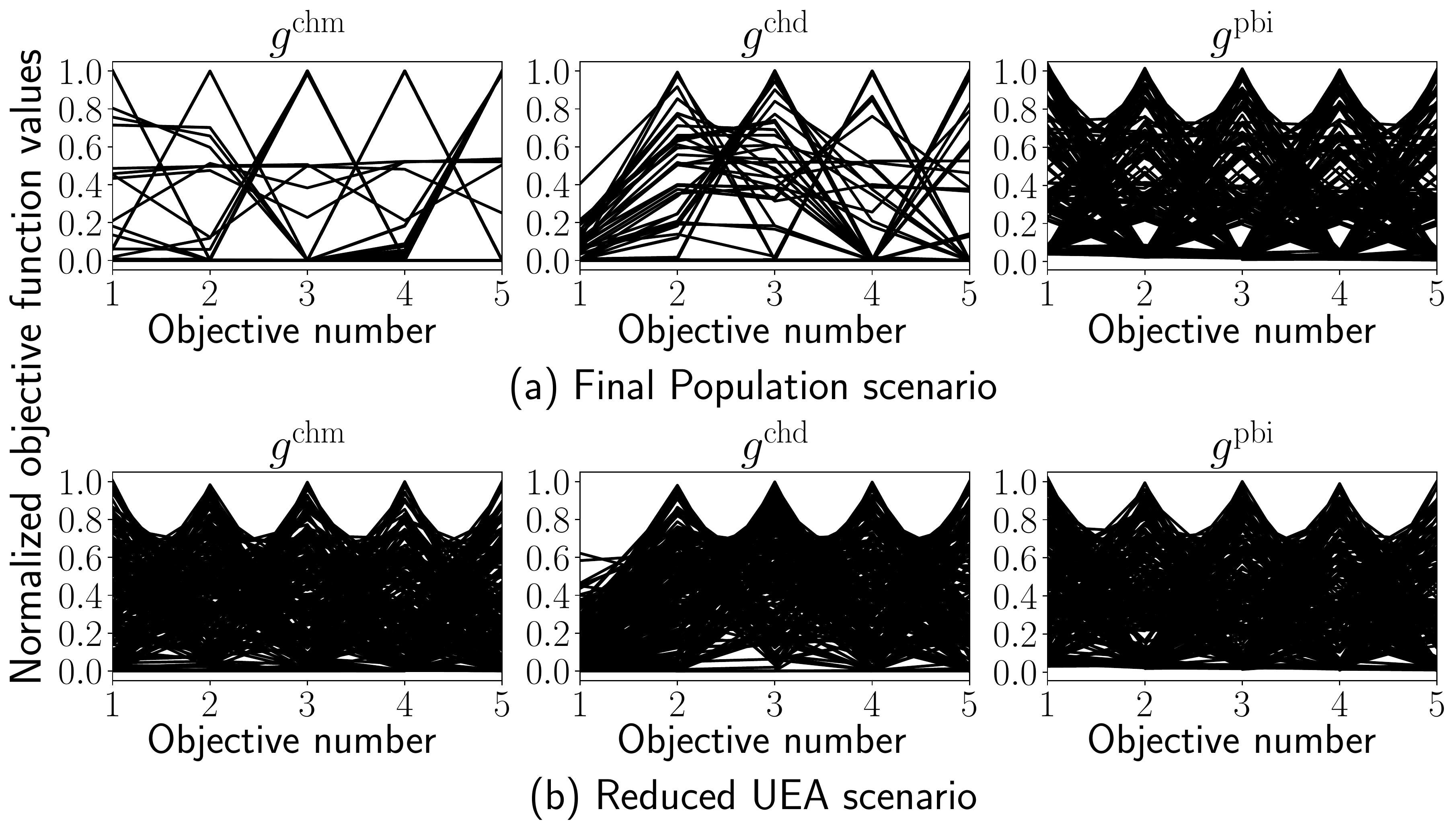}
\caption{
%
Parallel coordinates of the objective function values of nondominated solutions on the three-objective WFG4 problem for the final and reduced UEA scenarios.
Nondominated solutions shown here are found by MOEA/D with the three scalarizing functions ($g^{\rm chm}$, $g^{\rm chd}$, and $g^{\rm pbi}$ with $\theta=5$) at $5\times 10^4$ function evaluations.
The number of selected nondominated solutions $b$ for the reduced UEA scenario was set to $210$ for $M = 5$.
The horizontal and vertical axes represent the objective number and the normalized objective function values, respectively.
Results of a single run with the median HV value are shown.
}
\label{fig:sf_pc_wfg4}
\end{figure*}

\begin{figure*}[t]
\small
\newcommand{\widthvar}{0.99}
\centering
\includegraphics[width=\widthvar\textwidth]{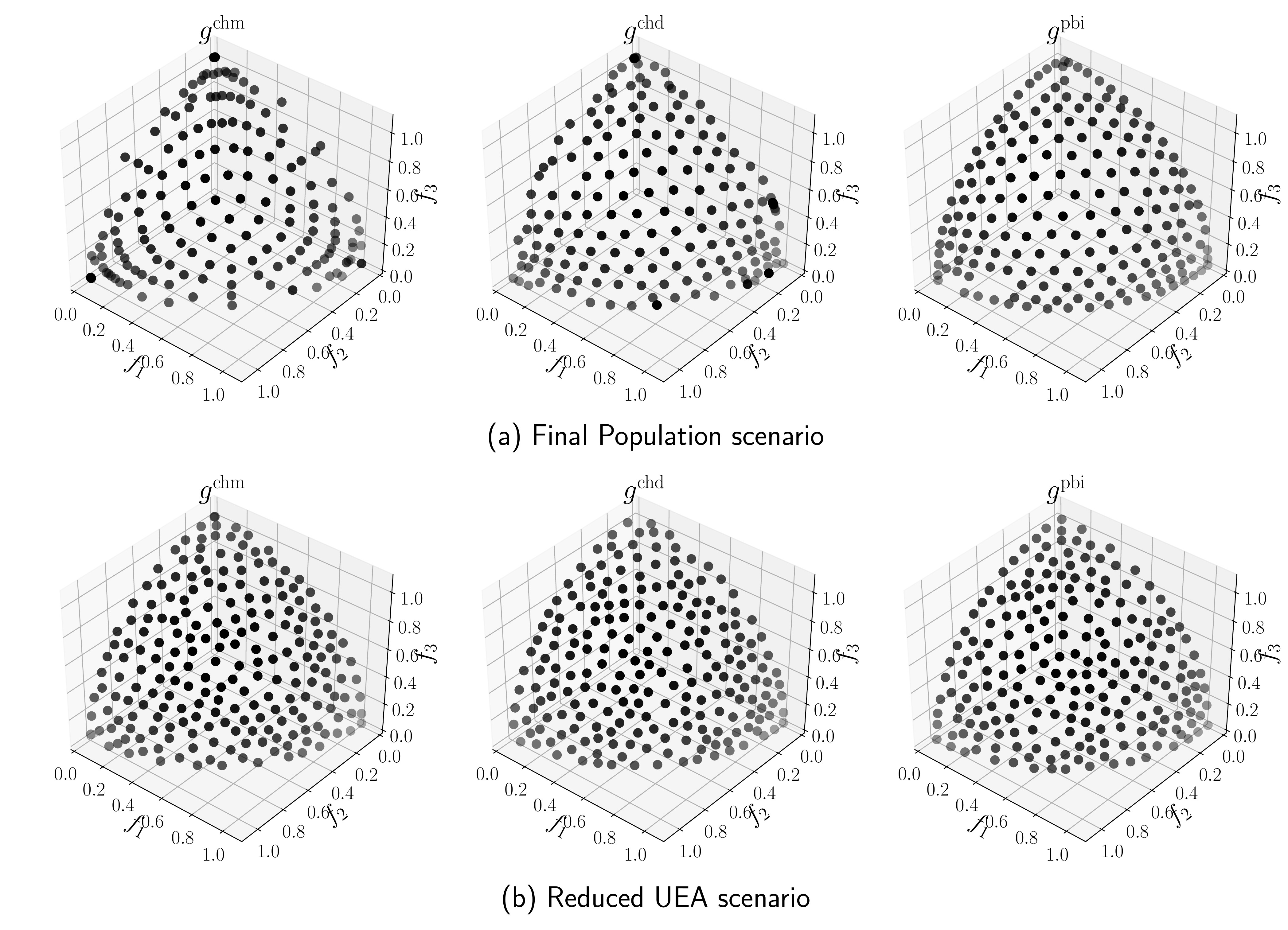}
\caption{
Distribution of nondominated solutions in the objective function space on the three-objective WFG4 problem for the final and reduced UEA scenarios.
Nondominated solutions shown here are found by MOEA/D with $g^{\rm chm}$, $g^{\rm chd}$, and $g^{\rm pbi}$ at $5\times 10^4$ function evaluations.
The number of selected nondominated solutions $b$ for the reduced UEA scenario was set to $210$ for $M = 3$.
Results of a single run with the median HV value are shown.
}
\label{fig:sf_3d_distribution_wfg4}
\end{figure*}


Figure \ref{fig:sf_pc_wfg4} shows the parallel coordinates of the objective function values of nondominated solutions found by MOEA/D with the three scalarizing functions on the five-objective WFG4 problem under (a) the final population scenario and (b) the reduced UEA scenario.
It should be recalled that $g^{\rm pbi}$ and $g^{\rm chm}$ are the best scalarizing functions on the five-objective WFG4 problem for the final population and reduced UEA scenarios, respectively (see Figure \ref{fig:sf_wfg4}).

The number of nondominated solutions in Figure \ref{fig:sf_pc_wfg4}(a) is $146$ for $g^{\rm chm}$, $150$ for  $g^{\rm chd}$, and $210$ for $g^{\rm pbi}$.
That is, MOEA/D with $g^{\rm pbi}$ obtains the largest number of nondominated solutions among the three configurations for the final population scenario.
The nondominated solutions found by MOEA/D with $g^{\rm pbi}$ are also uniformly distributed in the objective function space.
For these two reasons, it seems that MOEA/D with $g^{\rm pbi}$ achieves the highest HV value at the end of the search.
In contrast to the results for the final population scenario, as shown in Figure \ref{fig:sf_pc_wfg4}(b), MOEA/D with $g^{\rm chm}$ obtains well-distributed nondominated solutions in the objective function space under the reduced UEA scenario.
The uniformity of the distribution of nondominated solutions found by MOEA/D with $g^{\rm chm}$ is slightly better than those obtained by the other methods.

For further discussion, we show the distribution of nondominated solutions in the objective function space for both the final and reduced UEA scenarios on the three-objective WFG4 problem in Figure \ref{fig:sf_3d_distribution_wfg4}.
The distribution on the three-objective WFG4 problem in Figure \ref{fig:sf_3d_distribution_wfg4} is similar to the distribution on the five-objective WFG4 problem in Figure \ref{fig:sf_pc_wfg4}.
In the final population scenario (Figure \ref{fig:sf_3d_distribution_wfg4}(a)), while the uniformly distributed nondominated solutions are obtained by using $g^{\rm pbi}$ and $g^{\rm chd}$, the distribution of solutions found by MOEA/D with $g^{\rm chm}$ is biased.
However, it seems that MOEA/D with all the three scalarizing functions find evenly distributed nondominated solution under the reduced UEA scenario (Figure \ref{fig:sf_3d_distribution_wfg4}(b)).

As pointed out in \cite{TanabeIO17,LiYL16}, it seems that MOEA/D with $g^{\rm chm}$ is capable of generating well-approximated solutions but cannot maintain them in its population.
Such a problem of $g^{\rm chm}$ is addressed by incorporating the UEA, which automatically stores good nondominated solutions independently from the population, into MOEA/D.
Due to this reason, MOEA/D with $g^{\rm chm}$ performs the best on the five-objective WFG4 problem under the reduced UEA scenario, regarding the HV metric.
This observation can be found in some of the other test problems.


\subsection{Effect of the penalty value $\theta$ for the PBI function on the performance of MOEA/D}
\label{sec:results_pbi_theta}

\subsubsection{Overall performance}
\label{sec:results_pbi_theta_aps}

\begin{figure*}[t]
\small
\newcommand{\widthvar}{0.99}
\centering
\includegraphics[width=\widthvar\textwidth]{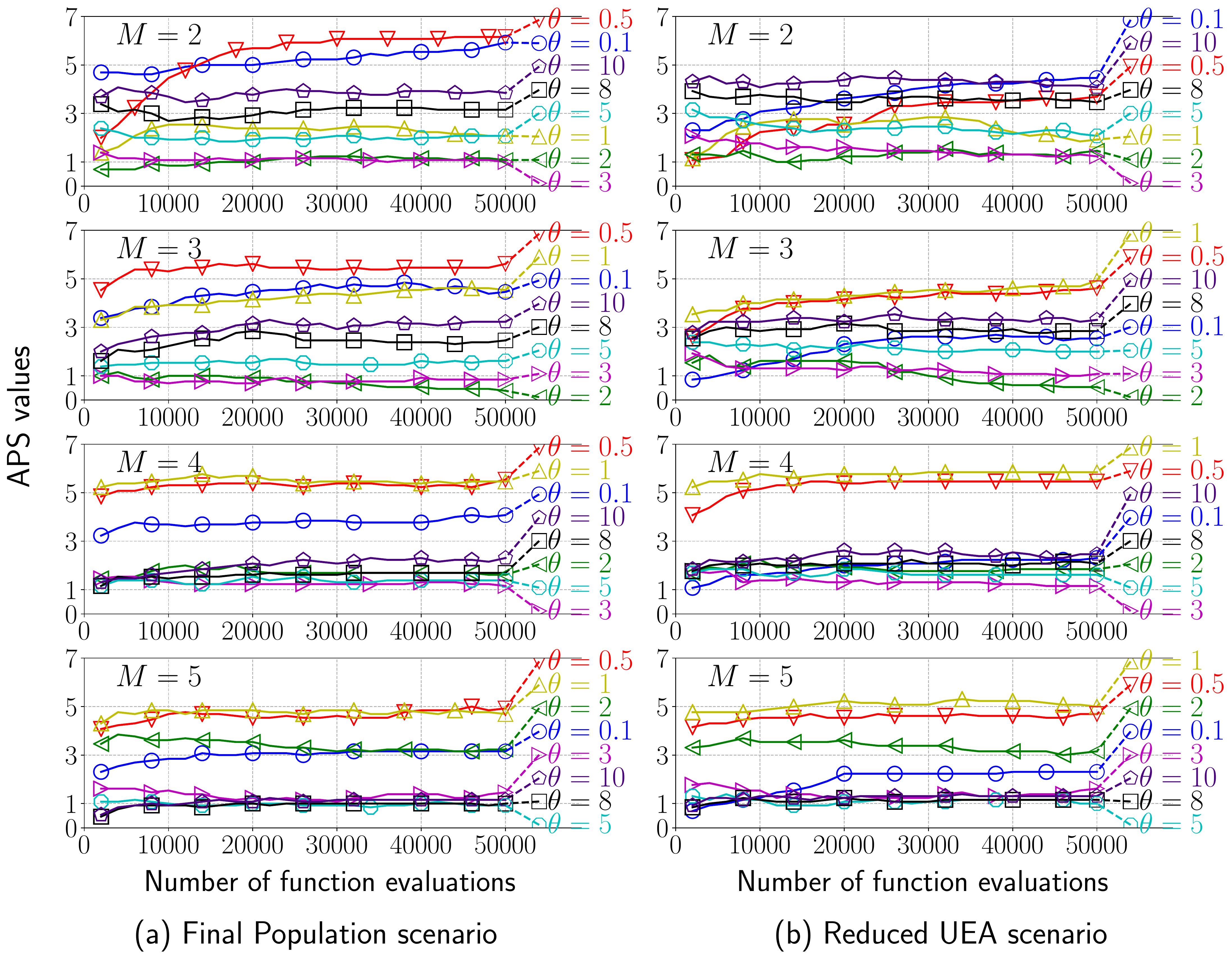}
\caption{
Overall performance of MOEA/D using the PBI function $g^{\rm pbi}$ with various $\theta$ values on the 13  problems with $M \in \{2, 3, 4, 5\}$.
The horizontal and vertical axes represent the number of function evaluations and the APS values, respectively.
For the APS value \cite{BaderZ11}, see Subsection \ref{sec:aps}.
}
\label{fig:pbitheta_aps}
\end{figure*}

Figure \ref{fig:pbitheta_aps} shows the overall performance of MOEA/D using the PBI function $g^{\rm pbi}$ with $\theta \in \{0.1, 0.5, 1, 2, 3, 5, 8, 10\}$ on the 13 problems with $M \in \{2, 3, 4, 5\}$ under  (a) the final population scenario and (b) the reduced UEA scenario.
Unlike the results of $\mu$ and $g$ presented in Subsections \ref{sec:results_mu} and \ref{sec:results_g}, the performance rank of each configuration is not drastically changed during the search process for both scenarios.
That is, a configuration that performs well at the beginning of the search exhibits good performance also at the end of the search.

As shown in Figure \ref{fig:pbitheta_aps}(a), good APS values are obtained by $\theta \in \{2, 3\}$ for $M \in \{2, 3\}$ for the final population scenario.
The best penalty value is $\theta \in \{3,5\}$ and $\theta \in \{5,8\}$ for $M=4$ and $M=5$, respectively.
It seems that MOEA/D using $g^{\rm pbi}$ with $\theta = 5$ works well for all $M$.
In contrast, $\theta \in \{0.1, 0.5, 1\}$ are clearly inappropriate penalty values for all $M$.

Figure \ref{fig:pbitheta_aps}(b) shows results for the reduced UEA scenario.
It seems that the performance rank under the reduced UEA scenario is not significantly different from that under the final population scenario.
Although the smallest APS value is obtained by $\theta = 0.1$ until $6 \, 000$ function evaluations for $M \geq 3$, MOEA/D using $g^{\rm pbi}$ with $\theta \in \{3, 5\}$ perform well among all the configurations for all $M$.
While the performance rank of $\theta = 0.1$ is improved $M \geq 3$, MOEA/D using $g^{\rm pbi}$ with $\theta \in \{0.5, 1\}$ performs poorly even for the reduced UEA scenario.

\begin{figure*}[t]
\small
\newcommand{\widthvar}{0.99}
\centering
\includegraphics[width=\widthvar\textwidth]{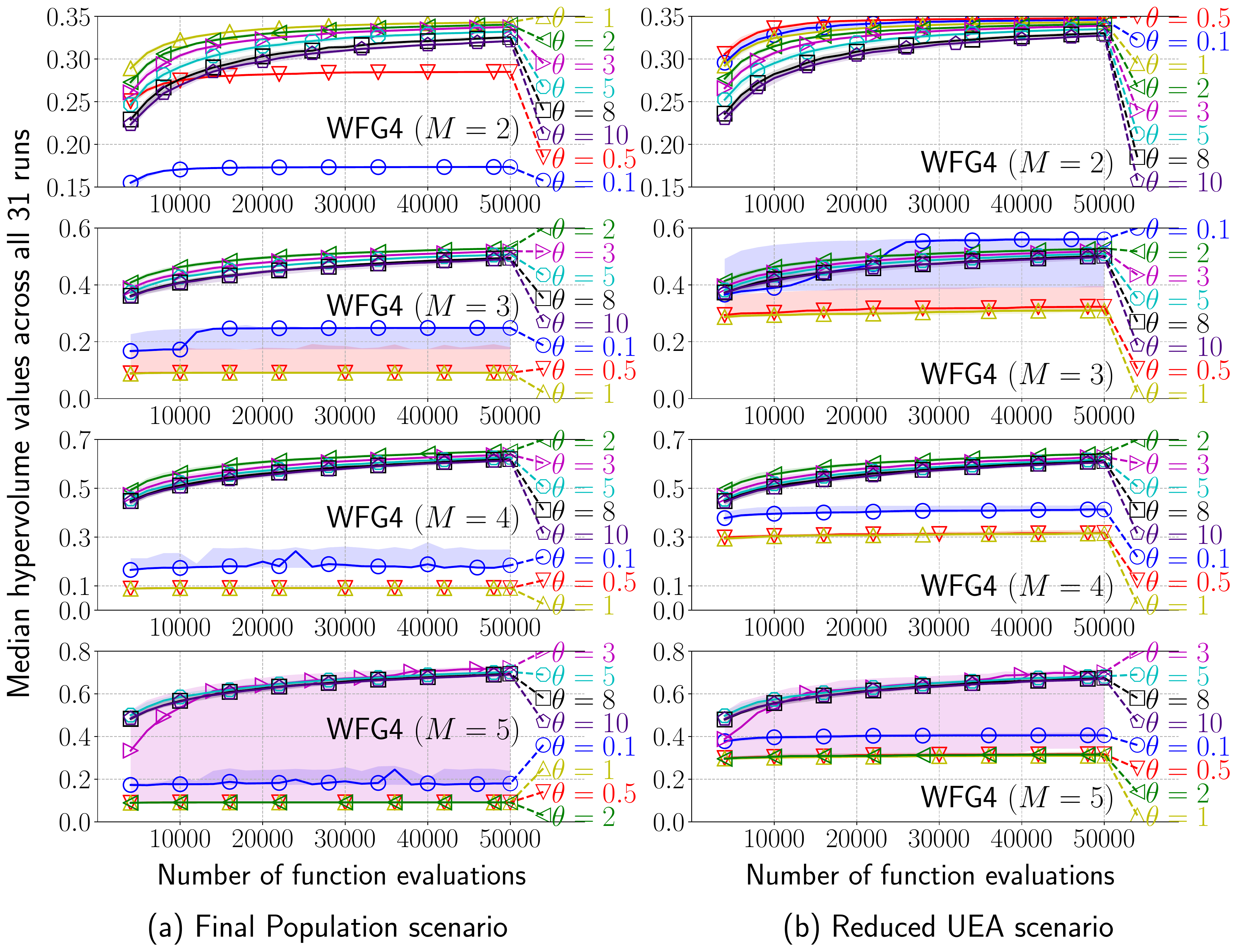}
\caption{
Performance of MOEA/D using the PBI function $g^{\rm pbi}$ with various $\theta$ values on the WFG4 problem with $M \in \{2, 3, 4, 5\}$.
The horizontal and vertical axes represent the number of function evaluations and the HV values, respectively.
The shaded area indicates 25-75 percentiles.
}
\label{fig:pbitheta_wfg4}
\end{figure*}

\subsubsection{Results on each problem}
\label{sec:results_pbi_theta_each_problem}

Figure \ref{fig:pbitheta_wfg4} shows the anytime performance of MOEA/D using $g^{\rm pbi}$ with various $\theta$ values on the WFG4 problem with $M \in \{2, 3, 4, 5\}$.
Figures S.27 -- S.39 in the supplementary file show results on all the 13 problems.

As seen from Figure \ref{fig:pbitheta_wfg4}(a), for the final population scenario, $\theta = 1, 2, 2$, and $3$ are suitable values on the WFG4 problem with $M=2, 3, 4,$ and $5$, respectively.
That is, the best penalty value $\theta$ increases with respect to the number objectives $M$ on the WFG4 problem.
Interestingly, as shown in Figure \ref{fig:pbitheta_wfg4}(b), in contrast to the results for the final population scenario, MOEA/D using $g^{\rm pbi}$ with $\theta = 0.1$ achieves the good HV value under the reduced UEA scenario for $M \in \{2, 3\}$.
%
%
According to the percentiles in Figure \ref{fig:pbitheta_wfg4}(a) and (b), for $M=3$, the variation of the HV values obtained by using $\theta \in \{0.1, 0.5\}$ is larger than that by using other $\theta$ values.
Interestingly, in Figure \ref{fig:pbitheta_wfg4}, the 25th and 75th percentile values are not evenly distributed.
For example, on the results for $M=3$ in Figure \ref{fig:pbitheta_wfg4}(a), while the median and the 25th percentile values are almost the same, the difference between the median and 75th percentile values is large.
Only for $M=5$, a large variation is observed in results of MOEA/D with $\theta = 3$.
While MOEA/D with $\theta = 3$ obtains the best HV value under both scenarios regarding the median HV value, it performs similarly to MOEA/D with $\theta \in \{0.1, 0.5, 1, 2\}$ for a particular run as seen from its 75th percentile values.

Due to space constraints, results on other test problems are not described here.
Similar to the results on the WFG4 problem, the best $\theta$ value significantly depends on the considered scenario on most test problems.


\begin{figure*}[t]
\small
\newcommand{\widthvar}{0.99}
\centering
\includegraphics[width=\widthvar\textwidth]{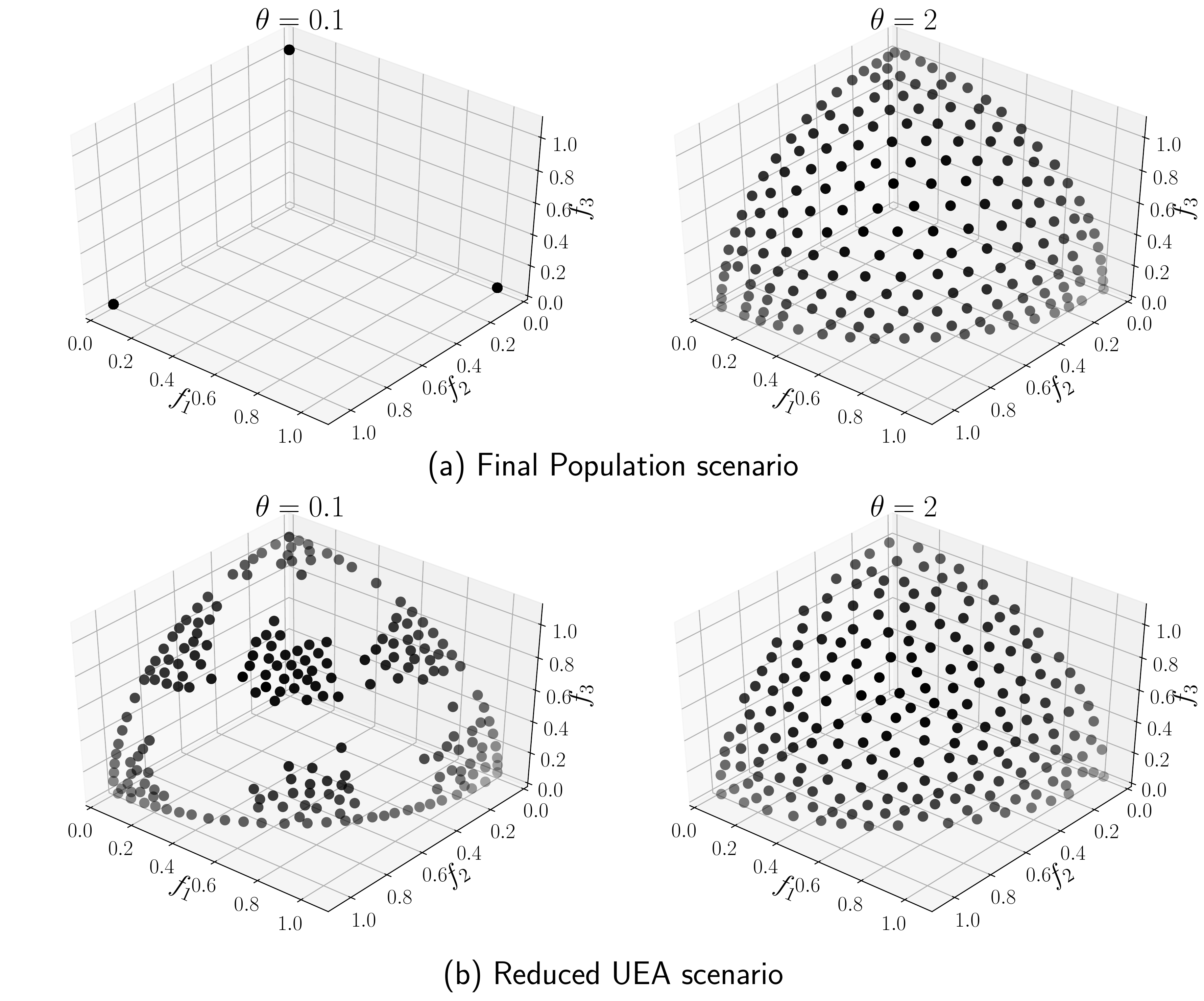}
\caption{
Distribution of nondominated solutions in the objective function space on the three-objective WFG4 problem for the final and reduced UEA scenarios.
Nondominated solutions shown here are found by MOEA/D using the PBI function $g^{\rm pbi}$ with $\theta = 0.1$ (left) and  with $\theta = 2$ (right) at $5\times 10^4$ function evaluations.  
Results of a single run with the median HV value are shown.
}
\label{fig:pbitheta_distribution_wfg4}
\end{figure*}

\subsubsection{Discussion}
\label{sec:results_pbi_theta_discussion}

Our results described in Subsections \ref{sec:results_pbi_theta_aps} and \ref{sec:results_pbi_theta_each_problem} show that the best setting of $\theta$ is significantly dependent on the characteristics of a given problem. 
For example, while MOEA/D using $g^{\rm pbi}$ with $\theta =2$ performs well on the WFG4 problem with $M=3$ (Figure \ref{fig:pbitheta_wfg4}(a)),  $\theta =10$  is the best setting for solving the three-objective DTLZ1 problem (Figure S.27).
Such conclusions are consistent with previous studies \cite{IshibuchiAN15,MohammadiOLD15,IshibuchiDN16,Sato15}.
%
For example, experimental results in \cite{IshibuchiAN15} show that MOEA/D using $g^{\rm pbi}$  with $\theta = 0.1$ performs better than that with other $\theta$ values on multi- and many-objective knapsack problems with the convex PFs.
However, a small penalty value like $\theta = 0.1$ is not appropriate for MOEA/D to solve MOPs with nonconvex PFs \cite{IshibuchiDN16}.

In addition to the  characteristics of a given problem, our results in Subsections \ref{sec:results_pbi_theta_aps} and \ref{sec:results_pbi_theta_each_problem} newly reveal that a suitable $\theta$ value also depends on a target optimization scenario.
For example, as shown in Figure \ref{fig:pbitheta_wfg4}, the best $\theta$ value is totally different on the WFG4 problem (nonconvex) for each scenario.
Below, we discuss the reason why the  $\theta = 0.1$ is the best penalty value on the three-objective WFG4 problem for the reduced UEA scenario.


Figure \ref{fig:pbitheta_distribution_wfg4} shows the distribution of nondominated solutions found by MOEA/D using $g^{\rm pbi}$ with $\theta = 0.1$ and $\theta = 2$ on the three-objective WFG4 problem for the final and reduced UEA scenarios.
It should be noted that $\theta = 2$ is the best parameter setting on the WFG4 problem with $M=3$ for the final population scenario (see Figure \ref{fig:pbitheta_wfg4}).
As seen from Figure \ref{fig:pbitheta_distribution_wfg4}(a), while MOEA/D with $\theta = 0.1$ can find only the three extreme solutions under the final population scenario, well-distributed nondominated solutions are obtained by $\theta = 2$.
However, Figure \ref{fig:pbitheta_distribution_wfg4}(b) shows that in addition to the extreme solutions, MOEA/D with $\theta = 0.1$ achieves some nondominated solutions distributed on the rim and inside regions of the PF.
Note that the HV value of nondominated solutions obtained by using $\theta = 0.1$ is higher than that by $\theta = 2$  in Figure \ref{fig:pbitheta_distribution_wfg4}(b).

It is pointed out that the HV indicator does not always assess the uniformity and diversity of the distribution of nondominated solutions \cite{IshibuchiISN17}.
For further investigation, we used the inverted generational distance (IGD) \cite{ZitzlerTLFF03}, which evaluates both convergence and diversity performance of MOEAs.
The generational distance (GD) and maximum spread (MS) \cite{ZitzlerTLFF03} indicator values of nondominated solutions are also reported to discuss convergence and diversity performance of MOEA/D, individually.

Below, let $\vector{A}$ be a set of nondominated solutions selected from the population or the UEA.
A set of reference vectors are denoted as $\vector{A}^*$, where each element of $\vector{A}^*$ is a Pareto-optimal solution.
The GD value is the average distance from each solution in $\vector{A}$ to its nearest reference vector in $\vector{A}^*$ in the objective function space:
%
 \begin{align}
\label{eqn:gd}
{\rm GD} (\vector{A}) &= \frac{1}{|\vector{A}|} \left(\sum_{\vector{x} \in \vector{A}} \min_{\vector{z} \in \vector{A}^*} \Bigl\{ {\rm ED} \bigl(\vector{f}(\vector{x}), \vector{f}(\vector{z}) \bigr) \Bigr\} \right),
 \end{align}
%
\noindent where ${\rm ED}(\vector{x}, \vector{y})$ is the Euclidean distance between $\vector{x}$ and $\vector{y}$.
In contrast to the GD metric in equation \eqref{eqn:gd}, the IGD value is the average distance from each reference vector in $\vector{A}^*$ to its nearest solution in $\vector{A}$ in the objective function space as follows:
%
  \begin{align}
\label{eqn:igd}
{\rm IGD} (\vector{A}) &= \frac{1}{|\vector{A}^*|} \left(\sum_{\vector{z} \in \vector{A}^*} \min_{\vector{x} \in \vector{A}} \Bigl\{ {\rm ED} \bigl(\vector{f}(\vector{x}), \vector{f}(\vector{z}) \bigr) \Bigr\} \right).
 \end{align}

The MS value of $\vector{A}$ is the average of the maximum and minimum objective function values of solutions in $\vector{A}$ for each objective as follows:
%
  \begin{align}
\label{eqn:ms}
{\rm MS} (\vector{A}) &= \sqrt{\sum^M_{i=1} \Bigl(\max_{\vector{x} \in \vector{A}}  \{f_i(\vector{x}) \} - \min_{\vector{x} \in \vector{A}}  \{f_i(\vector{x}) \}\Bigr)^2 }.
 \end{align}

\begin{figure*}[t]
\small
\newcommand{\widthvar}{0.99}
\centering
\includegraphics[width=\widthvar\textwidth]{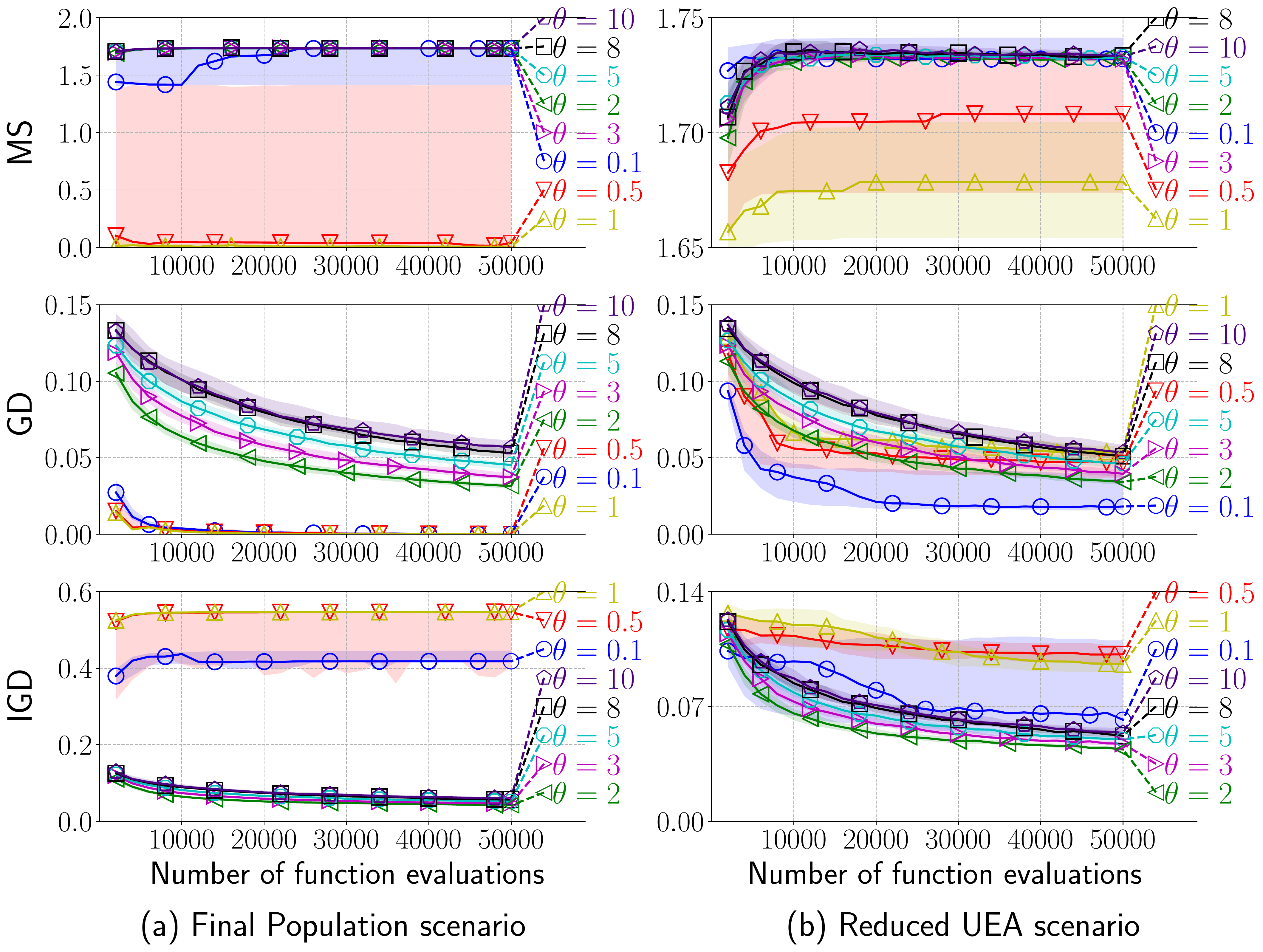}
\caption{
The MS, GD, and IGD indicator values of nondominated solutions found by MOEA/D using $g^{\rm pbi}$ with various $\theta$ values on the three-objective WFG4 problem under (a) the final population scenario and (b) the reduced UEA scenario.
A low GD value and a high MS value indicate that the corresponding method has good convergence and diversity performance, respectively.
A method showing a low IGD value performs well in terms of both convergence and diversity.
The shaded area indicates 25-75 percentiles.
}
\label{fig:pbitheta_wfg4_msgdigd}
\end{figure*}

Figure \ref{fig:pbitheta_wfg4_msgdigd} shows the MS, GD, and IGD indicator values of nondominated solutions found by MOEA/D using $g^{\rm pbi}$ with various $\theta$ values on the three-objective WFG4 problem under (a) the final population scenario and (b) the reduced UEA scenario.
Before calculating each indicator value, all objective function values were normalized as described in Subsection \ref{sec:problems}.
One may wonder that GD values for the final population scenario are significantly better than those for the reduced UEA scenario.
This is because the number of nondominated solutions in the population is usually smaller than that in the UEA (e.g., see Figure \ref{fig:pbitheta_distribution_wfg4}), and a small number of solutions is possibly beneficial for GD as discussed in \cite{IshibuchiMN15}.

As seen from Figure \ref{fig:pbitheta_wfg4_msgdigd}, MOEA/D with $\theta \in \{0.1, 2, 3, 5, 8, 10\}$ achieve high MS values for both scenarios.
The best GD values are also obtained by using $\theta = 0.1$ for the reduced UEA scenario.
Thus, nondominated solutions in the UEA obtained by using $\theta = 0.1$ are close to the PF.
This is because the contour lines of $g^{\rm pbi}$ with $\theta = 0.1$ and the weighted sum function are similar \cite{IshibuchiAN15}.
Therefore, MOEA/D using $g^{\rm pbi}$ with $\theta = 0.1$ shows good HV values on the WFG4 problem with $M=3$.

It is interesting to notice that the performance rank based on the IGD metric is slightly different from that based on the HV indicator.
Although the best HV values are obtained by using $\theta = 0.1$ under the reduced UEA scenario (Figure \ref{fig:pbitheta_wfg4}), MOEA/D with $\theta = 2$ achieves the lowest IGD values (Figure \ref{fig:pbitheta_wfg4_msgdigd}).
Since the IGD metric assesses the uniformity of the distribution, the IGD value of nondominated solutions found by MOEA/D with $\theta = 0.1$, whose distribution is biased to specific regions as shown in Figure \ref{fig:pbitheta_distribution_wfg4}, is worse than that by MOEA/D with $\theta = 2$.

\section{Further analysis}
\label{sec:further_discussion}



On the one hand, in Section \ref{sec:experimental_results}, we examined the influence of each control parameter on MOEA/D by keeping other parameters default (see Table \ref{tab:moead_parameter_settings}).
This is because we wanted to examine the effect of changing each parameter in a component-wise manner.
If two or more parameters are perturbed simultaneously, which parameter improves or degrades the performance of MOEA/D may be unclear.

On the other hand, analysis of dependencies between multiple control parameters is important and interesting.
It is expected that some control parameters (e.g., $\mu$ and $g$) are correlated with each other.
Also, we used the default parameters (see Table \ref{tab:moead_parameter_settings}), which are commonly used in the literature, for our analysis.
However, the effect of the three control parameters ($\mu$, $g$, and $\theta$) on MOEA/D may be different when well-tuned parameters are used instead of the default parameters.
Thus, whether our analysis results in Section \ref{sec:experimental_results} are representative or not is not guaranteed.


This section presents further analysis of MOEA/D in order to address the above-mentioned issues.
Dependencies between two control parameters are examined in Subsection \ref{sec:correlation_analysis}.
Analysis of each control parameter with the best configuration is presented in Subsection \ref{sec:best_config}.

\begin{figure*}[t]
\small
\newcommand{\widthvar}{0.99}
\centering
\includegraphics[width=\widthvar\textwidth]{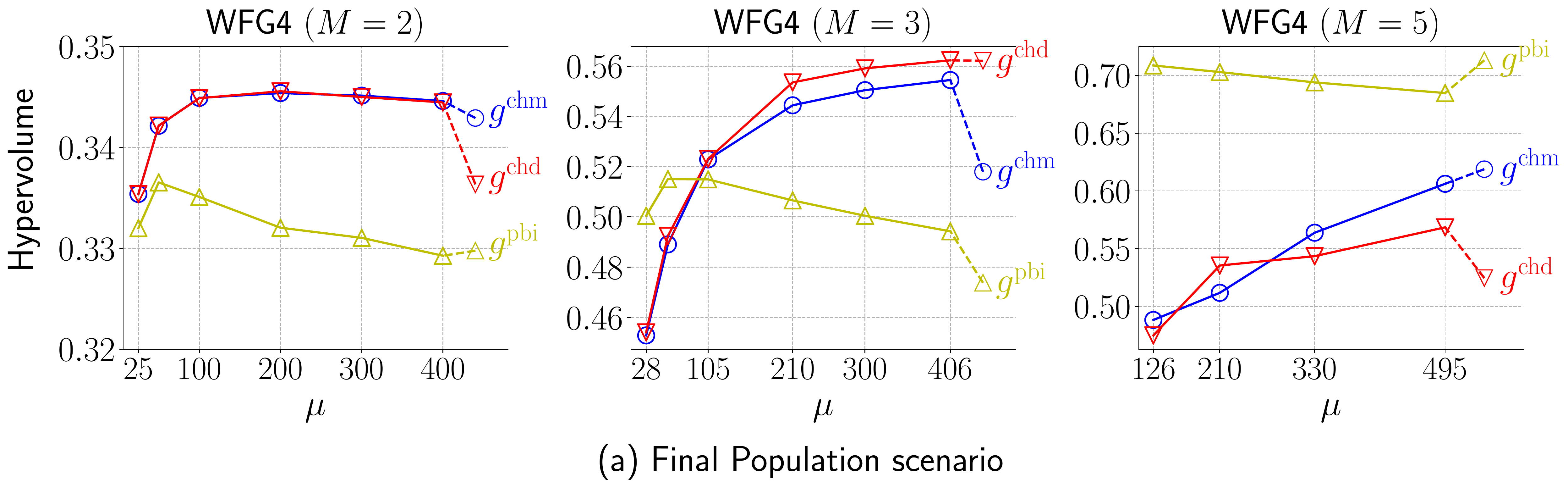}
\\
\includegraphics[width=\widthvar\textwidth]{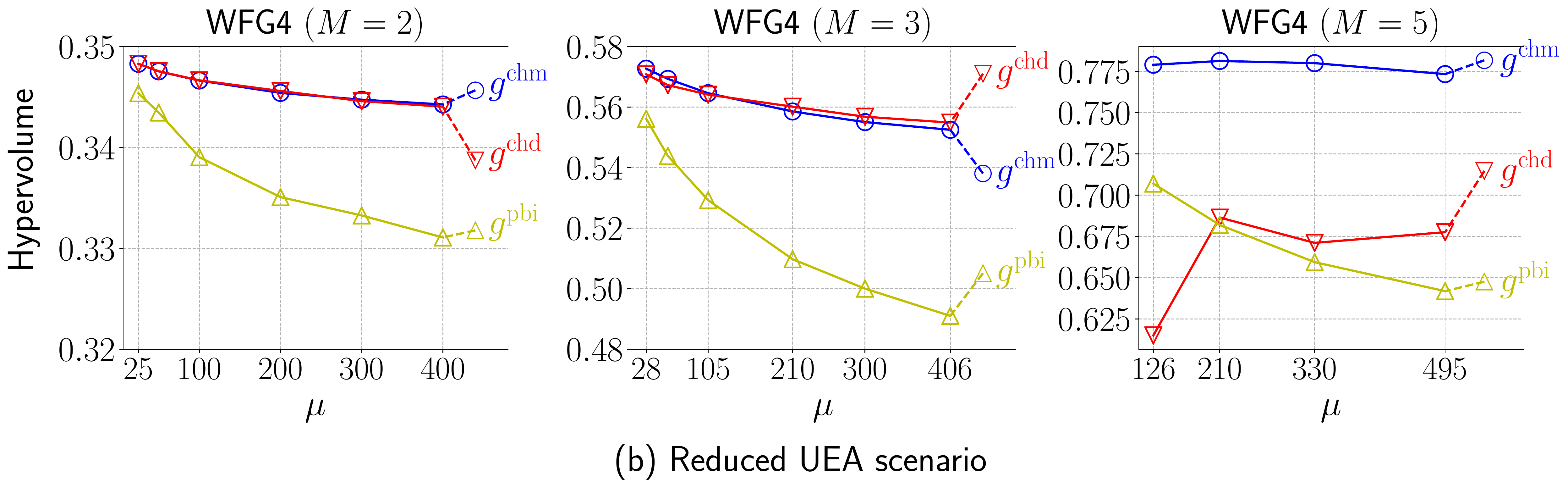}
\caption{
Influence of $\mu$ on the performance of MOEA/D with the three scalarizing functions ($g^{\rm chm}$, $g^{\rm chd}$, and $g^{\rm pbi}$) on the WFG4 problem with $M \in \{2, 3, 5\}$.
The median HV value at $50\,000$ evaluations among 31 runs is shown.
}
\label{fig:mu_vs_sf}
\end{figure*}

\begin{figure*}[t]
\small
\newcommand{\widthvar}{0.99}
\centering
\includegraphics[width=\widthvar\textwidth]{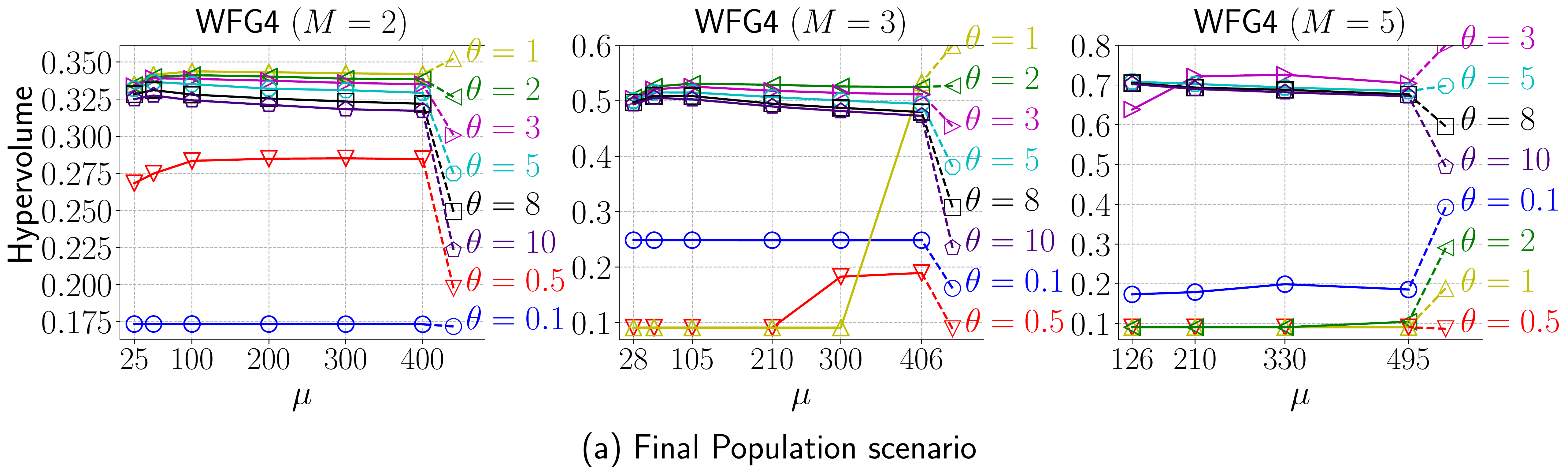}
\includegraphics[width=\widthvar\textwidth]{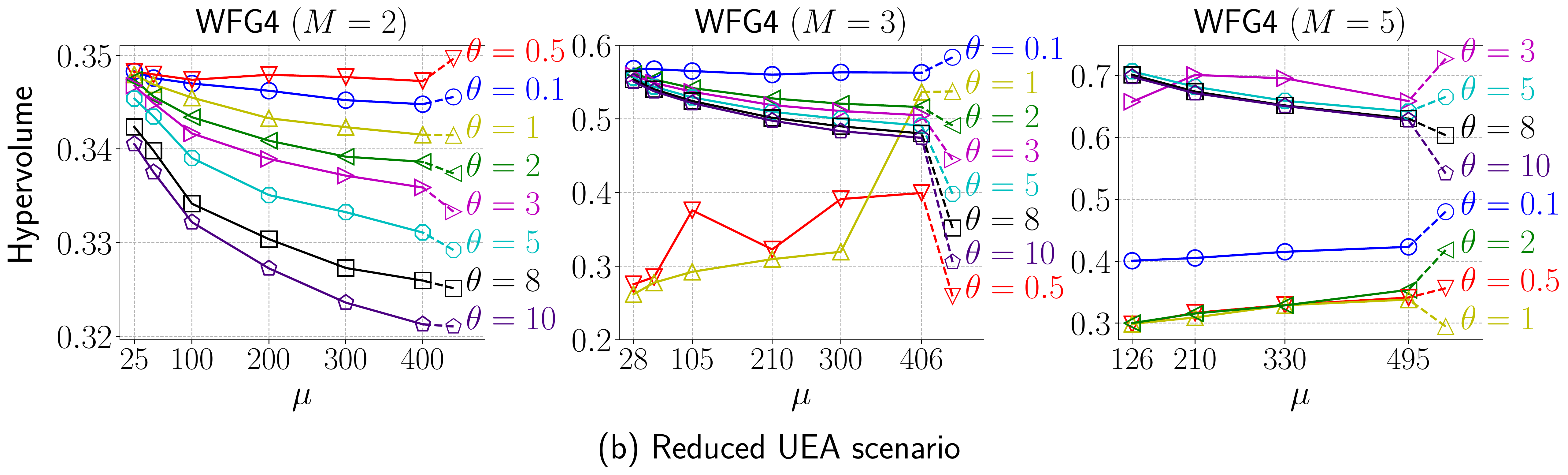}
\caption{
Influence of $\mu$ on the performance of MOEA/D using $g^{\rm pbi}$ with various $\theta$ values on the WFG4 problem with $M \in \{2, 3, 5\}$.
The median HV value at $50\,000$ evaluations among 31 runs is shown.
}
\label{fig:mu_vs_theta}
\end{figure*}

\begin{figure*}[t]
\small
\newcommand{\widthvar}{0.99}
\centering
\includegraphics[width=\widthvar\textwidth]{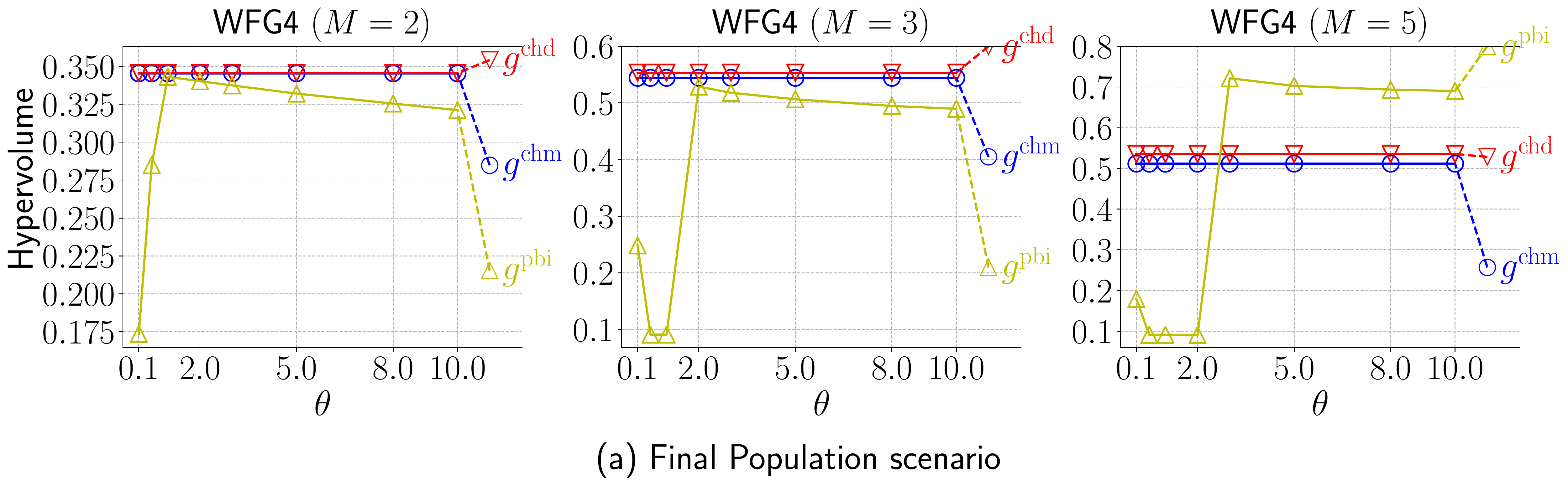}
\\
\includegraphics[width=\widthvar\textwidth]{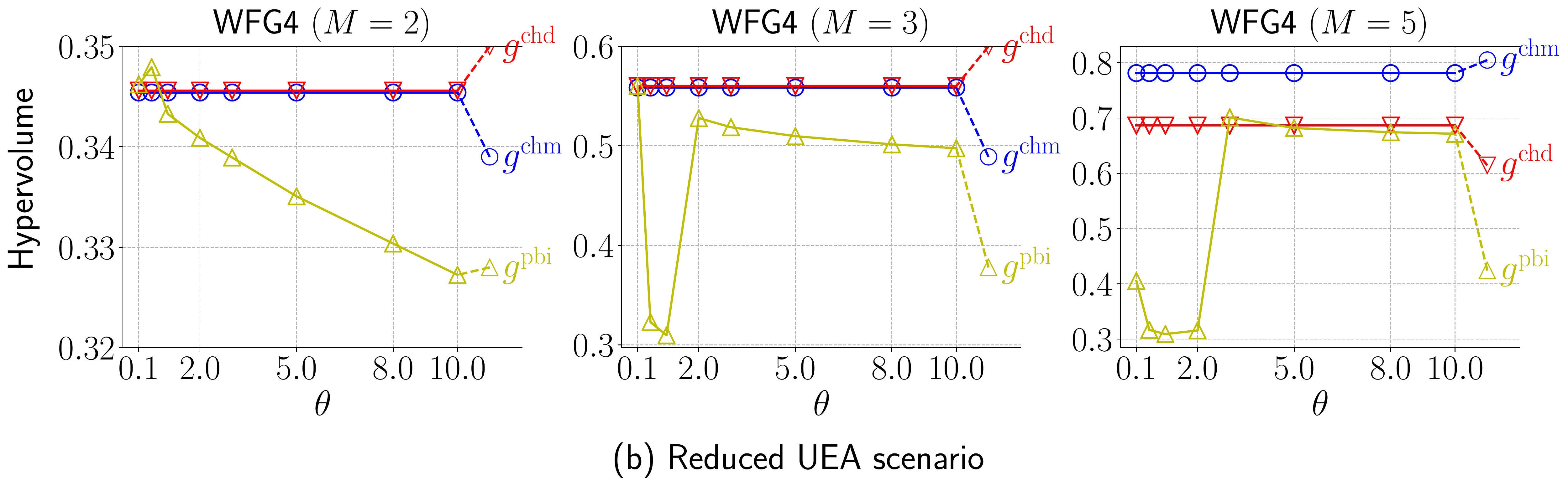}
\caption{
Comparison of the two Chebyshev functions ($g^{\rm chm}$ and $g^{\rm chd}$) and $g^{\rm pbi}$ with various $\theta$ values on the WFG4 problem with $M \in \{2, 3, 5\}$.
The median HV value at $50\,000$ evaluations among 31 runs is shown.
}
\label{fig:theta_vs_sf}
\end{figure*}

\subsection{Analysis of dependencies between two control parameters}
\label{sec:correlation_analysis}

While we discussed the results based on the anytime performance of MOEA/D in Section \ref{sec:experimental_results}, we report only the end-of-the-run results in this subsection.
Although the end-of-the-run results do not provide sufficient information as described in Section \ref{sec:experimental_results}, analyzing dependencies between two parameters based on the anytime performance is not a trivial task.
Such an analysis is our future research.


\subsubsection{Dependencies between $\mu$ and $g$}

Figure \ref{fig:mu_vs_sf} shows the influence of $\mu$ on the performance of MOEA/D with the three scalarizing functions ($g^{\rm chm}$, $g^{\rm chd}$, and $g^{\rm pbi}$) on the WFG4 problem with $M \in \{2, 3, 5\}$.
Results on other test problems can be found in Figures S.40 -- S.52 in the supplementary file.

As seen from Figure \ref{fig:mu_vs_sf}, the best $\mu$ value is different depending on scalarizing functions.
For example, as shown in Figure \ref{fig:mu_vs_sf}(a), for the final population scenario, MOEA/D with the two Chebyshev functions need a relatively large $\mu$ value for all $M$.
In contrast, small $\mu$ values are suitable for $g^{\rm pbi}$.
In addition, as discussed in Subsection \ref{sec:results_mu}, suitable $\mu$ values also depend on the choice of an optimization scenario.
Figure \ref{fig:mu_vs_sf}(b) indicates that small $\mu$ values are appropriate for all the three scalarizing functions on the WFG4 problem under the reduced UEA scenario.
For $M \in \{2, 3\}$, in most cases, the increase of $\mu$ deteriorates the performance of MOEA/D under the reduced UEA scenario.
In summary, $\mu$ and $g$ are correlated with each other, and the relation between them is dependent on the choice of a scenario.

\subsubsection{Dependencies between $\mu$ and $\theta$}

Figure \ref{fig:mu_vs_theta} shows the impact of $\mu$ on the performance of MOEA/D using $g^{\rm pbi}$ with various $\theta$ values on the WFG4 problem with $M \in \{2, 3, 5\}$.
Results on other test problems are shown in Figures S.53 -- S.65 in the supplementary file.
Unlike the analysis results of $\mu$ and $g$, the effect of $\mu$ values on the performance of MOEA/D with various $\theta$ values is relatively moderate, except for some results (e.g., results for $\theta = 1$ on the three-objective WFG4 problem).
Thus, the performance rank of MOEA/D with different $\theta$ values is not significantly influenced by $\mu$.

\subsubsection{Comparison of the two Chebyshev functions and the PBI function with various $\theta$ values}

Figure \ref{fig:theta_vs_sf} provides a comparison of the two Chebyshev functions ($g^{\rm chm}$ and $g^{\rm chd}$) and $g^{\rm pbi}$ with $\theta \in \{0.1, ..., 10\}$ on the WFG4 problem with $M \in \{2, 3, 5\}$.
For $g^{\rm chm}$ and $g^{\rm chd}$, the same HV values are shown for each $\theta$ value in Figure \ref{fig:theta_vs_sf}.
Results on other test problems can be found Figures S.79 -- S.91 in the supplementary file.

As seen from Figure \ref{fig:theta_vs_sf}, the performance of MOEA/D with $g^{\rm pbi}$ significantly depends on the $\theta$ value for both optimization scenarios.
Although such results under the final population scenario have already been reported in previous work (e.g., \cite{MohammadiOLD15,Sato15}), similar results are observed under the reduced UEA scenario.
MOEA/D with a particular $\theta$ value performs better than that with the two Chebyshev functions.
Whereas the poor results are obtained by the $g^{\rm pbi}$ with $\theta = 5$ under the reduced UEA scenario in Subsection \ref{sec:results_g}, it is likely that MOEA/D using $g^{\rm pbi}$ with a suitable $\theta$ value performs well.
Interestingly, the performance of MOEA/D regarding the HV metric is drastically improved or degraded at around $\theta \in \{1, 2, 3\}$.
The reason is unclear, and thus its further analysis is our future research topic.






\subsection{Analysis using well-tuned parameters}
\label{sec:best_config}

In Section \ref{sec:experimental_results}, except for a single parameter to be examined, other parameters were set to the default values as shown in Table \ref{tab:moead_parameter_settings}.
In contrast, here, we examine the effect of each parameter on MOEA/D by using the best parameters.

We conducted a grid search to find the best configuration for each control parameter.
For the three control parameters ($\mu$, $g$, and $\theta$), all the values in Table \ref{tab:moead_parameter_settings} were examined.
In addition to the three control parameters ($\mu$, $g$, and $\theta$), the performance of MOEA/D with four neighborhood sizes $T \in \{10, 20, 40, 80\}$ was investigated.
We evaluated the performance of MOEA/D with all possible combinations of the four parameters ($240$ and $160$ configurations for $M \in \{2, 3, 4\}$ and $M=5$ respectively) on each problem instance according to the same experimental procedure described in Section \ref{sec:experimental_settings}.
The best configuration for each parameter was determined according to the best median HV value at the $50\,000$ evaluations on each test problem.
Due to space constraints, the best parameters are not shown, but they are different depending on each control parameter and each problem instance.

\begin{figure*}[t]
\small
\newcommand{\widthvar}{0.99}
\centering
\includegraphics[width=\widthvar\textwidth]{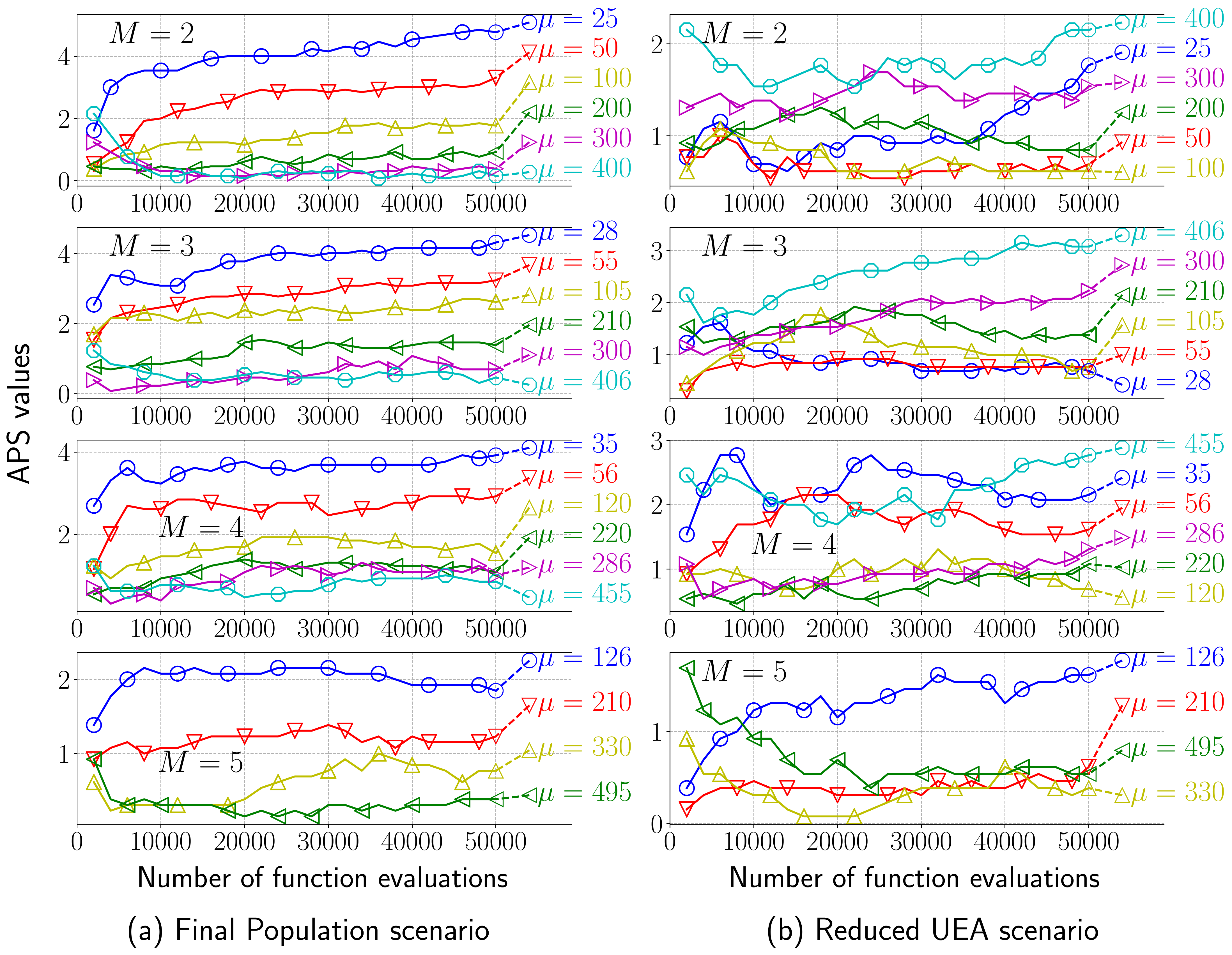}
\caption{
Overall performance of MOEA/D with various $\mu$ settings on the 13 problems with $M \in \{2, 3, 4, 5\}$ (lower is better).
The horizontal and vertical axes represent the number of function evaluations and the APS values, respectively.
The best configurations ($g$, $\theta$, and $T$) on each problem instance are used for each $\mu$ value (i.e., the control parameters used are different for different problems).
}
\label{fig:popsize_aps_bestconfig}
\end{figure*}

\begin{figure*}[t]
\small
\newcommand{\widthvar}{0.99}
\centering
\includegraphics[width=\widthvar\textwidth]{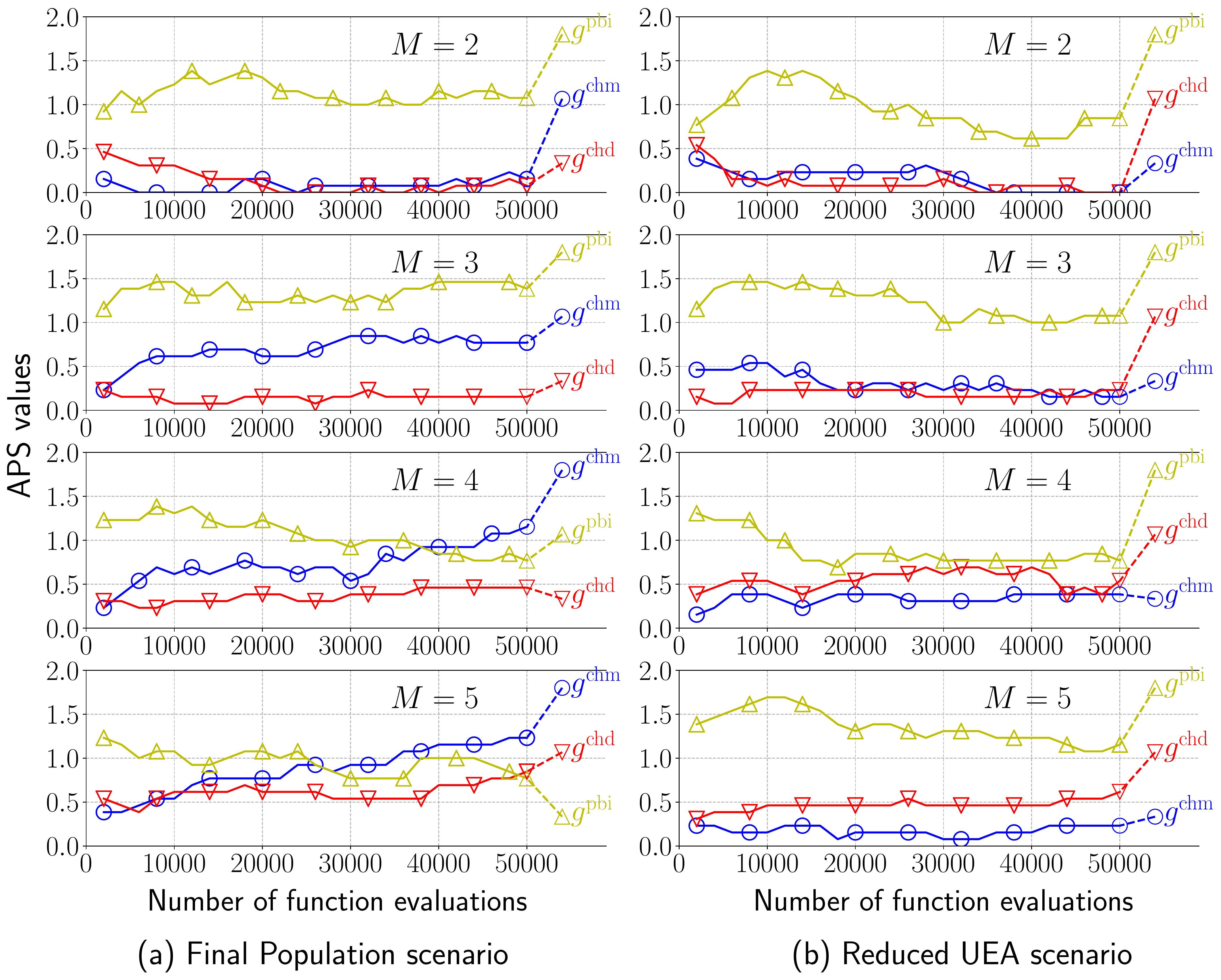}
\caption{
Overall performance of MOEA/D with the three scalarizing functions ($g^{\rm chm}$, $g^{\rm chd}$, and $g^{\rm pbi}$) on the 13 problems with $M \in \{2, 3, 4, 5\}$ (lower is better).
The horizontal and vertical axes represent the number of function evaluations and the APS values, respectively.
The best configurations ($\mu$, $\theta$, and $T$) on each problem instance are used for each scalarizing function.
}
\label{fig:sf_aps_bestconfig}
\end{figure*}

\begin{figure*}[t]
\small
\newcommand{\widthvar}{0.99}
\centering
\includegraphics[width=\widthvar\textwidth]{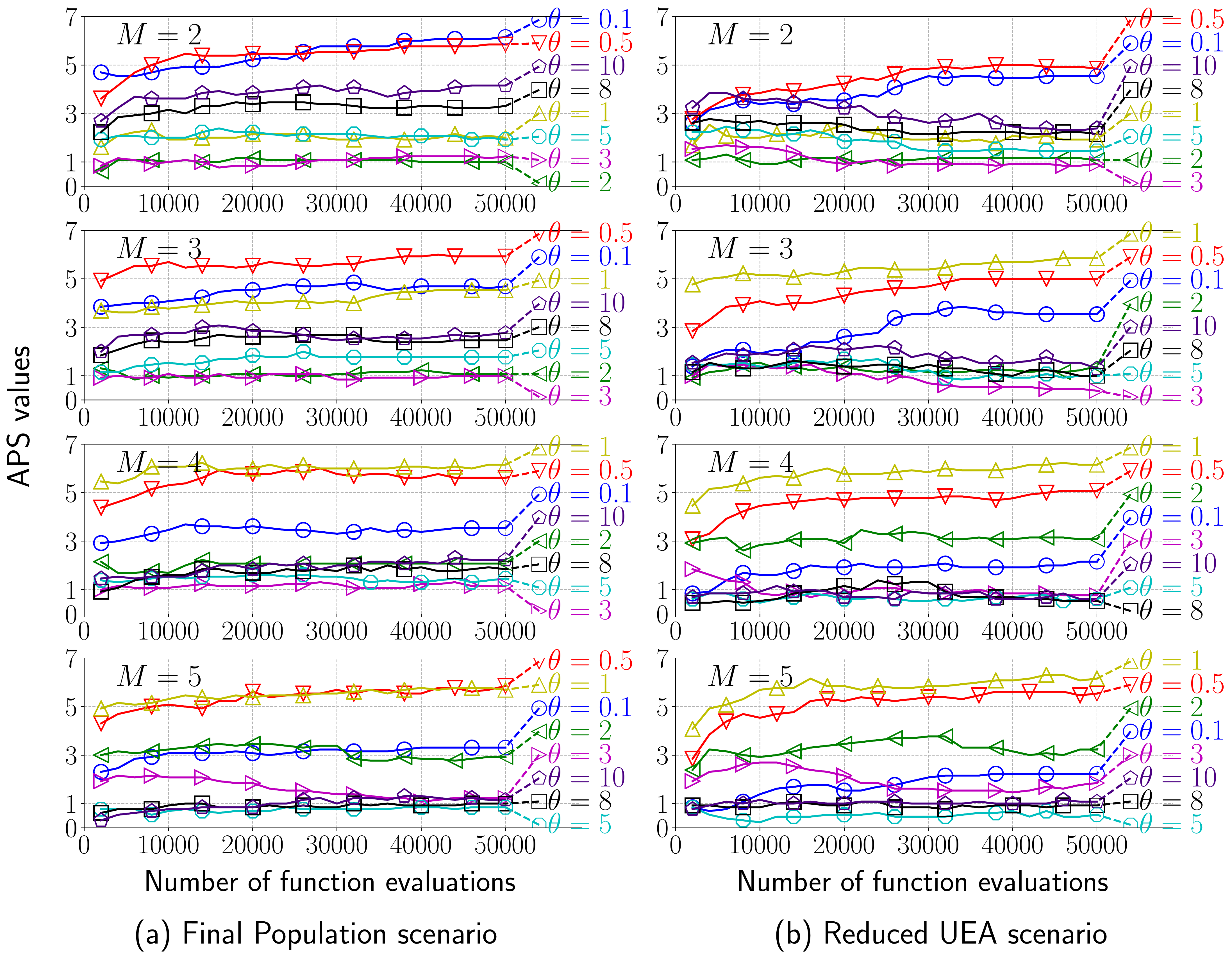}
\caption{
Overall performance of MOEA/D using the PBI function $g^{\rm pbi}$ with various $\theta$ values on the 13  problems with $M \in \{2, 3, 4, 5\}$.
The horizontal and vertical axes represent the number of function evaluations and the APS values, respectively.
The best configurations ($\mu$ and $T$) on each problem instance are used for each $\theta$ value.
}
\label{fig:pbitheta_aps_bestconfig}
\end{figure*}

Figures \ref{fig:popsize_aps_bestconfig}, \ref{fig:sf_aps_bestconfig}, and \ref{fig:pbitheta_aps_bestconfig} show the influence of $\mu$, $g$, and $\theta$ on the overall performance of MOEA/D with the best configurations on the 13 test problems, respectively.
For each problem instance, for each optimization scenario, the best configuration for each control parameter to be analyzed was used.
%
%
The performance rank of MOEA/D with the best parameters (Figures \ref{fig:popsize_aps_bestconfig}, \ref{fig:sf_aps_bestconfig}, and \ref{fig:pbitheta_aps_bestconfig}) is slightly different from that with the default parameters (Figures \ref{fig:popsize_aps}, \ref{fig:sf_aps}, and \ref{fig:pbitheta_aps}).
For example, unlike the results of MOEA/D with the default parameters (Figure \ref{fig:popsize_aps}), MOEA/D with $\mu = 400$ performs the best on the two-objective problems under the final population scenario (Figure \ref{fig:popsize_aps_bestconfig}).
The interval which MOEA/D with $g^{\rm pbi}$ performs the best for $M=5$ under the final population scenario becomes shorter (see Figures \ref{fig:sf_aps}(a) and \ref{fig:sf_aps_bestconfig}(a)).
Although $\theta=3$ is the second best $\theta$ value for $M=3$ under the reduced UEA scenario (Figure \ref{fig:pbitheta_aps}), $\theta=3$ is the most suitable for MOEA/D (Figure \ref{fig:pbitheta_aps_bestconfig}).

However, a qualitative difference between the results of MOEA/D with the default and best parameters is not significant.
Thus, the effect of the three control parameters on MOEA/D is almost the same even if the best parameters are used instead of the default parameters.
This result guarantees that our analysis results in Section \ref{sec:experimental_results} can be generalized to a certain degree.


\section{Conclusion}
\label{sec:conclusion}




In this paper, we analyzed the influence of the three control parameters (the population size $\mu$, the scalarizing function $g$, and the penalty value $\theta$ of the PBI function $g^{\rm pbi}$) on the performance of MOEA/D under the two different optimization scenarios (the final population and reduced UEA scenarios).
Our results on the four DTLZ and nine WFG problems up to five objectives show that a suitable setting of the three control parameters ($\mu$, $g$, and $\theta$) of MOEA/D is totally different depending on the problem and the optimization scenario.
Our observations are summarized as follows:
\begin{itemize}
\item While MOEA/D with a small $\mu$ value exhibits poor performance measured by the HV indicator on most of the 13 MOPs under the final population scenario, it performs significantly better than MOEA/D with a large $\mu$ value under the reduced UEA scenario.
\item For the final population scenario, the PBI function $g^{\rm pbi}$ with $\theta = 5$ is an appropriate scalarizing function for some MOPs, and such a conclusion is consistent with previous studies.
However,  for the reduced UEA scenario, MOEA/D with the two Chebyshev functions ($g^{\rm chm}$ and $g^{\rm chd}$) perform significantly better than that with $g^{\rm pbi}$ with $\theta = 5$.  
\item  While well-distributed nondominated solutions are obtained by only using a large $\theta$ value under the final population scenario, MOEA/D with a small $\theta$ value (e.g., $\theta = 0.1$) performs well on some MOPs under the reduced UEA scenario.
\end{itemize}

We also analyzed the reason why an appropriate control parameter setting is different for each scenario. 
Our analytical results can be briefly summarized as follows:
\begin{itemize}
\item MOEA/D with a particular parameter setting (e.g., a small $\mu$ value, $g^{\rm chm}$, and $g^{\rm pbi}$ with a small $\theta$ value) is capable of generating good nondominated solutions but cannot keep them in the population.
\item Such an issue can be addressed by incorporating the UEA into MOEA/D, which maintains well-distributed nondominated solutions independently from the population.
\item Therefore, an MOEA/D configuration that performs poorly for the final population scenario is probable to show a good performance for the reduced UEA scenario.
\end{itemize}

%

Although in this paper we presented the analysis of the three control parameters ($\mu$, $g$, and $\theta$), latest MOEA/D-type algorithms (e.g., MOEA/D-DE \cite{LiZ09}) have other control parameters, such as the neighborhood size $T$, the number of replacement individuals $n^{\rm rep}$, and the probability of selecting types of neighborhood $\delta$.
An analysis of remaining control parameters of MOEA/D-type algorithms is an avenue for future work.


 \section*{Acknowledgement}

This work was supported by the Science and Technology Innovation Committee Foundation of Shenzhen (Grant No. ZDSYS201703031748284).

\section*{References}

\bibliography{reference}

\end{document}